%% file: main.tex
\documentclass{article}

\usepackage{arxiv}

\usepackage[utf8]{inputenc}
\usepackage[T1]{fontenc}

\usepackage{amsmath}
\usepackage{amssymb}
\usepackage{amsfonts}

\usepackage{graphicx}
\usepackage{booktabs}
\usepackage{multirow}
\usepackage{wrapfig}
\usepackage{microtype}
\usepackage{placeins}

\usepackage[ruled,vlined]{algorithm2e}

\usepackage{tikz}
\usetikzlibrary{
    arrows.meta,
    positioning,
    fit,
    calc
}

\usepackage{url}
\usepackage{hyperref}

\hypersetup{
    colorlinks=true,
    linkcolor=blue,
    citecolor=blue,
    urlcolor=blue
}

\newcommand{\best}[1]{\textbf{#1}}
\newcommand{\second}[1]{\underline{#1}}

\date{}

\title{
OSOR: One-Step Diffusion Inpainting for Effect-Aware Object Removal
}

\author{
\mbox{Qinming~Zhou\textsuperscript{1,2,*}}
\quad
\mbox{Chenxi~Sun\textsuperscript{1,3,*}}
\quad
\mbox{Deyang~Kong\textsuperscript{1,3}}
\quad
\mbox{Junhao~He\textsuperscript{1}}
\\[0.3em]
\mbox{Xiangheng~Tang\textsuperscript{1,4}}
\quad
\mbox{Peike~Yu\textsuperscript{1,5}}
\quad
\mbox{Haotian~Wu\textsuperscript{1}}
\quad
\mbox{Leilei~Cao\textsuperscript{6,\(\dagger\)}}
\quad
\mbox{Linfeng~Zhang\textsuperscript{1,\(\ddagger\)}}
\\[0.8em]
{\normalfont\small
\textsuperscript{1}Shanghai Jiao Tong University
\quad
\textsuperscript{2}Tsinghua University
}
\\[0.2em]
{\normalfont\small
\textsuperscript{4}Xidian University
\quad
\textsuperscript{5}Tongji University
}
\\[0.2em]
{\normalfont\small
\textsuperscript{3}University of Electronic Science and Technology of China
\quad
\textsuperscript{6}Transsion
}
\\[0.5em]
{\normalfont\footnotesize
\textsuperscript{*}Equal contribution.
\quad
\textsuperscript{\(\dagger\)}Project Leader.
\quad
\textsuperscript{\(\ddagger\)}Corresponding Author.
}
\\[0.5em]
{\normalfont\ttfamily\small
\href{https://github.com/Zhouqm-Git/osor}{https://github.com/Zhouqm-Git/osor}
}
}

\begin{document}

\maketitle
\vspace{-0.5em}

\begin{figure}[htbp]
    \vspace{-0.8em}
    \centering
    \includegraphics[width=0.9\linewidth]{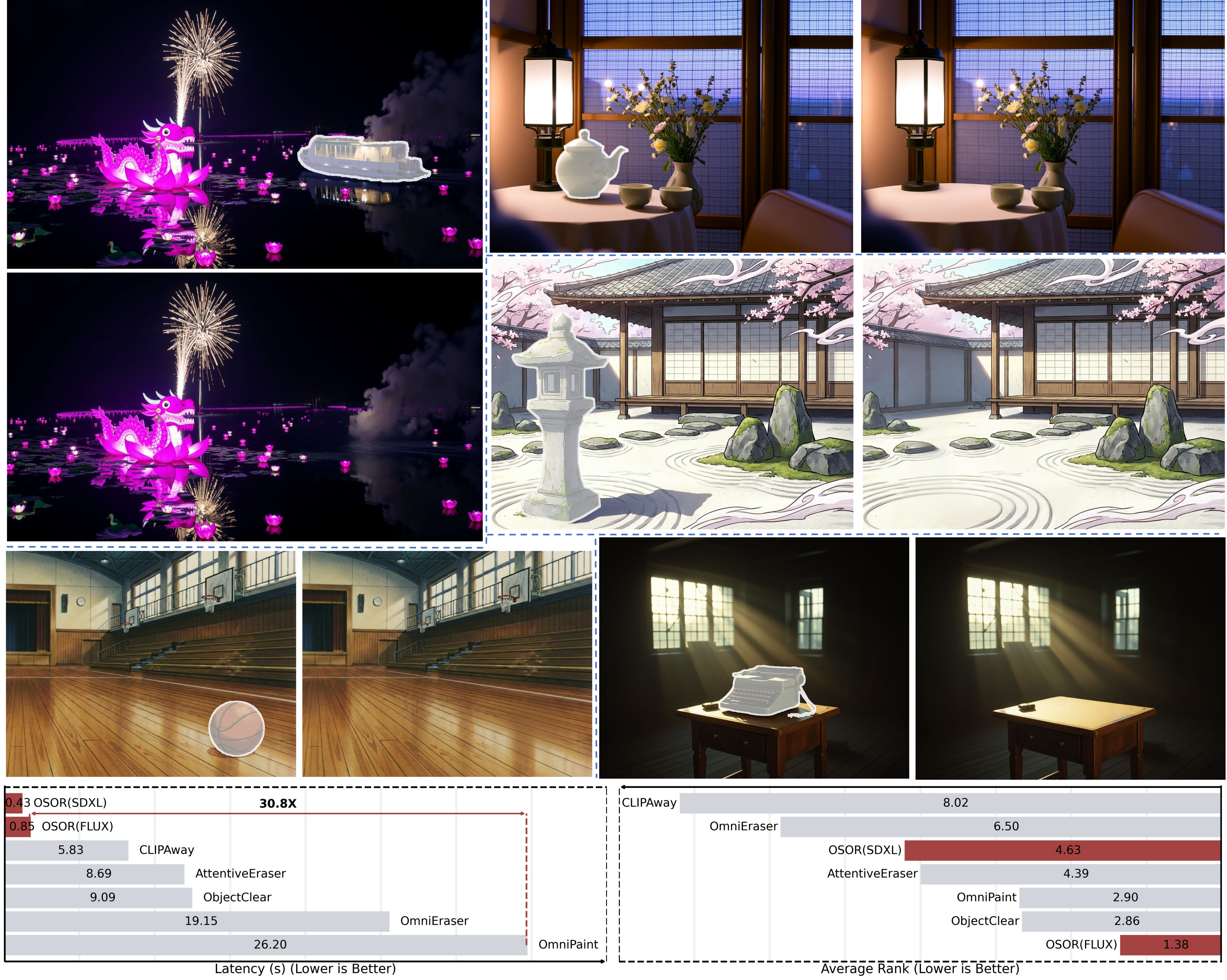}
    \caption{\textbf{Comparison between OSOR and other methods.}
    OSOR effectively removes object-associated effects, such as shadows,
    while running $10.6\times$ faster than ObjectClear.
    A $1024\times1024$ image can be processed in under one second on a
    single NVIDIA A100 GPU.
    The average rank is computed across six benchmarks.}
    \label{fig:head}
\end{figure}

\begin{abstract}
Real-world object removal is challenging due to two key difficulties:
the target object's non-local effects, such as shadows and reflections,
which are difficult to model, and the fact that user-provided masks are
often inaccurate or incomplete.
With billions of parameters and tens of denoising steps, diffusion-based
models achieve strong removal performance at the expense of substantial
computational cost, limiting their use in interactive applications and
on edge devices.
To address these challenges, we present \textbf{OSOR}
(One-Step Object Removal), which simultaneously achieves efficient,
effect-aware, and mask-robust object removal.
Concretely, OSOR introduces:
(1) an occupancy-guided discriminator for precise boundary supervision,
enabling stable single-step diffusion training;
(2) an alpha head that leverages knowledge from pretrained diffusion
models to predict appropriate removal regions with minimal overhead,
thereby handling imperfect masks; and
(3) a semantic-anchored verification pipeline (SAVP) that filters noisy
instruction-based triplets to produce effect-aware supervision at scale.
Using SAVP, we curate \textbf{CORNE}, which contains 280K verified
removal pairs, and further annotate AnimeEraseBench and TextEraseBench
to evaluate performance on more complex removal tasks.
Experiments show that OSOR surpasses strong multi-step diffusion
baselines in perceptual quality while achieving
$4\times$ to $30\times$ faster inference.
Code and resources are available at \url{https://github.com/Zhouqm-Git/osor}.
\end{abstract}


\input{sec/2_intro}
\input{sec/3_related_work}
\input{sec/4_method}

\input{sec/5_exp}
\input{sec/6_conclusion}
\section*{Acknowledgements}
This work was supported by Transsion Holdings.

\bibliographystyle{splncs04}
\bibliography{main}

\clearpage

\appendix

\begin{center} 
    {\LARGE\bfseries Supplementary Material\par} 
\end{center}

\input{sec/7_supplementary}

\end{document}

%% file: sec/2_intro.tex
\section{Introduction}

Object removal is a fundamental image editing task that aims to eliminate a target object and its visual effects from a photograph, restoring a natural background in the affected region. Generally speaking, real-world object removal presents three major challenges as follows.

\noindent \textbf{(I) Effect-awareness.} Removing an object is not simply a matter of erasing the pixels within a specified mask. The target object often leaves behind persistent visual effects in the scene, including cast shadows, reflected appearances, and other environmental interactions. Eliminating these effects requires a strong semantic understanding of the scene geometry and lighting, as the model must infer what the background should look like without the object and its influence. 

\noindent  \textbf{(II) Mask-robustness.} In practical usage scenarios, the removal mask is typically provided by the user through interactive selection tools. These user-generated masks are frequently imperfect: they may be too large and cover unrelated content, or too small and fail to encompass the full extent of the object and its effects. This masks robustness issue is critical for deployment in real applications where users expect reasonable results even with careless input. 

\noindent  \textbf{(III) Efficiency.} Object removal is often performed on mobile devices in an interactive manner, introducing requirements on extremely low latency.

Early approaches to object removal predominantly employed Generative Adversarial Networks (GANs)~\cite{goodfellow2014gan,pathak2016context,iizuka2017globally,suvorov2022lama}. These methods leveraged adversarial learning where ground-truth removal results served as real samples and model outputs as fake samples. While GANs offered computational advantages through single forward passes in lightweight networks, their limited representational capacity constrained generation quality and prevented achieving satisfactory effect-aware and mask-robust results.
Recent advances in diffusion models have fundamentally transformed the landscape of image generation and editing tasks. Modern diffusion models leverage hundreds of billions of parameters and dozens of iterative denoising steps to achieve unprecedented generative capabilities. These powerful models have been successfully adapted to image inpainting and object removal tasks~\cite{ho2020ddpm,rombach2022ldm,lugmayr2022repaint,podell2023sdxl,blackforestlabs2024fluxfill}, demonstrating superior perceptual quality compared to traditional GAN-based approaches. Despite these improvements, diffusion-based methods face a critical bottleneck in computational efficiency. The multi-step denoising process requires substantial computing resources and inference time, making it impractical for deployment on edge devices or interactive applications where latency is critical. Furthermore, even with the powerful priors encoded in pretrained diffusion models, existing approaches still struggle to fully address the challenges of effect-aware and mask-robust removal~\cite{jiang2025smarteraser,zhao2025objectclear,wei2025omnieraser,ekin2024clipaway,sun2025attentiveeraser,yu2025omnipaint}. This indicates that simply using pretrained diffusion models is insufficient, and targeted training strategies are necessary to unlock their full potential for this specific task.

To solve this problem , in this paper, we propose OSOR, a \textbf{O}ne-\textbf{S}tep diffusion model for \textbf{O}bject \textbf{R}emoval that simultaneously achieves three goals: efficiency through single-step inference, effect-aware removal of shadows and reflections, and robustness to imperfect user masks. Our approach introduces three key technical contributions to address the aforementioned challenges.

\textbf{Occupancy-guided discriminator for single-step diffusion.} While adversarial learning based step distillation~\cite{yin2024dmd,sauer2024add} has been successfully utilized in image generation, we find that applying these methods directly to object removal leads to unsatisfactory results, particularly producing blurry boundaries around the removed object. This difficulty arises because single-step diffusion lacks the iterative refinement opportunity that allows multi-step methods to gradually correct errors in the generated content~\cite{zhao2025objectclear,sun2025attentiveeraser,blackforestlabs2024fluxfill}. To address this, we design an occupancy-guided discriminator that provides precise boundary supervision by computing fractional occupancy values at multiple scales from the input mask for each patch location. Additionally, we propose formulating object removal as a latent restoration task derived from image restoration principles, where weakly noising the input image reduces training difficulty~\cite{meng2021sdedit,lin2025hypir}.

\textbf{A lightweight alpha head  for imperfect removal mask correction.}  OSOR introduces an alpha head, which is implemented as a lightweight projection appended to the diffusion backbone. Leveraging the rich semantic knowledge already present in pretrained diffusion models~\cite{li2024drip,hu2024diffumatting,huang2024layerdiff}, this component can accurately recover the appropriate mask for removal with minimal additional parameters and computational overhead. Besides, we further propose a two-stage training curriculum: the model first learns the removal task with perfect masks, then adapts to handle imperfect masks to improve robustness.

\textbf{Semantic-anchored verification pipeline enabling effect-aware removal data at scale.} Based on existing image editing datasets, this pipeline combines semantic information with pixel-space gradient analysis to verify successful removal and detect the presence of object effects. This allows us to automatically generate substantial amounts of labeled data for training effect-aware removal models. Using this pipeline, we construct CORNE, a high-quality dataset containing 280K verified object removal pairs with effect-aware annotations~\cite{zhao2025objectclear,yu2025omnipaint}. Furthermore, to address the lack of comprehensive benchmarks for evaluating object removal in specific domains, we construct AnimeEraseBench and TextEraseBench, which evaluate removal capability on anime images and text overlays, respectively. In summary,
our contributions are threefold.
\begin{itemize}
    \item We propose OSOR, a single-step object removal model, which introduces an occupancy-guided discriminator, a lightweight alpha head, and a semantic-anchored verification pipeline (SAVP) to achieve efficient, effect-aware, and mask-robust object removal.
    \item Based on SAVP,  we curate CORNE, a high-quality and effect-aware object removal training dataset with 280K removal pairs. Besides, we release AnimeEraseBench and TextEraseBench, two removal benchmarks for evaluation on removal in anime scenarios and text objects, respectively.
    \item Extensive experimental results on 6 benchmarks with 7 comparison methods demonstrate the superior removal quality and efficiency. For instance, 2.24 dB higher PSNR and 27$\times$ faster than the second-best method in AnimeEraseBench.
\end{itemize}

%% file: sec/3_related_work.tex
\section{Related Work}

\subsection{Image Inpainting and Object Removal}
Object removal is related to mask-conditioned inpainting, but it imposes a stricter requirement to preserve surrounding context while removing the target object and its visual effects. GAN-based inpainting enables fast inference but often struggles with boundary continuity and high-frequency realism under irregular or large masks~\cite{pathak2016context,iizuka2017globally,yu2019freeform,suvorov2022lama}. Transformer-based inpainting improves global structure modeling for large missing regions~\cite{li2022mat,dong2022zits}. Diffusion-based inpainting further boosts realism by leveraging strong generative priors~\cite{rombach2022ldm,lugmayr2022repaint,avrahami2022blended,podell2023sdxl,blackforestlabs2024fluxfill}. However, general inpainting backbones are optimized for generic completion rather than removal behavior, so outputs remain sensitive to mask quality when object effects extend beyond the provided spatial conditioning.

\subsection{Object Removal Models and Datasets}
Recent works adapt diffusion models to object removal through stronger conditioning, task-oriented guidance, or effect-aware modeling~\cite{jiang2025smarteraser,zhao2025objectclear,wei2025omnieraser,liu2025erasediffusion,ekin2024clipaway,sun2025attentiveeraser,yu2025omnipaint}. Progress is also driven by improved supervision sources, including video-derived paired frames and captured counterfactual pairs that better reflect real effects~\cite{sagong2022rord,winter2024objectdrop,wei2025omnieraser}. In parallel, instruction-based corpora provide abundant editing triplets but can be noisy, motivating verification and filtering to obtain reliable supervision~\cite{brooks2022instructpix2pix,kuprashevich2025nohumansrequired}. A persistent difficulty in interactive removal is that user masks often under-cover soft effects. Several methods strengthen object effect coupling in conditioning or supervision, for example via object effect attention~\cite{zhao2025objectclear}. Another direction is to predict a soft effective editing region to represent uncertain boundaries and residual effects beyond the provided mask. This view is related to alpha compositing and matting, which model soft transitions with an opacity map instead of a hard mask~\cite{porter1984compositing,levin2008matting,xu2017deepmatting}. Recent diffusion-based matting further supports predicting alpha-like maps from diffusion priors as a representation of soft boundaries and layer uncertainty~\cite{li2024drip,hu2024diffumatting,huang2024layerdiff}.

\begin{figure}
\centering
\includegraphics[width=1\linewidth]{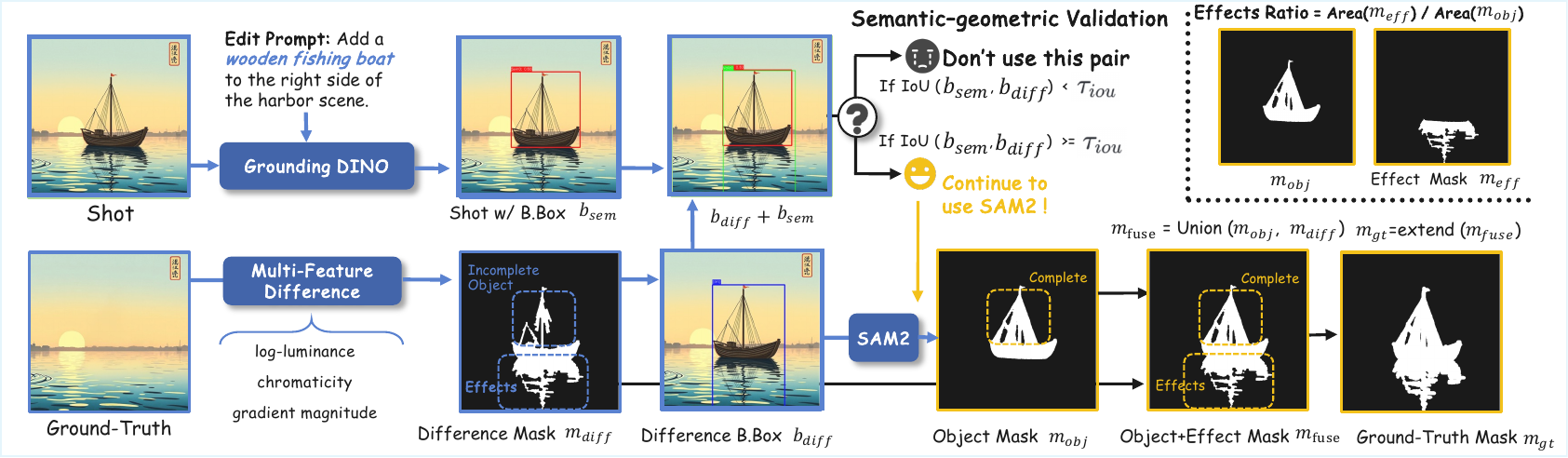}
\caption{\textbf{Overview of SAVP and CORNE.} Starting from single-edit instruction triplets, SAVP verifies semantically aligned and localized differences, then fuses the validated difference region with promptable segmentation to form an effect-aware mask. It further derives object-core masks for Phase~II incomplete-mask conditioning.}
\label{fig:dataset}
\end{figure}

\subsection{Efficient Diffusion and One-Step Generation}
Reducing diffusion inference cost has been studied through distillation and consistency training that reduces denoising steps~\cite{salimans2022progressive,song2023consistency,luo2023lcm}. A growing line of work targets one-step or near one-step generation, distilling pretrained diffusion models into single-pass generators via distribution matching and adversarial objectives~\cite{yin2024dmd,sauer2024add}. Most efficient diffusion methods are developed for global generation and do not explicitly enforce the strict context preservation required by mask-conditioned editing. Single-pass editing is also more sensitive to boundary ambiguity near the mask, which motivates task-specific supervision when applying one-step inference to object removal.

%% file: sec/4_method.tex
\section{Methodology}
\label{sec:method}

\begin{figure}
    \centering
    \includegraphics[width=0.95\linewidth]{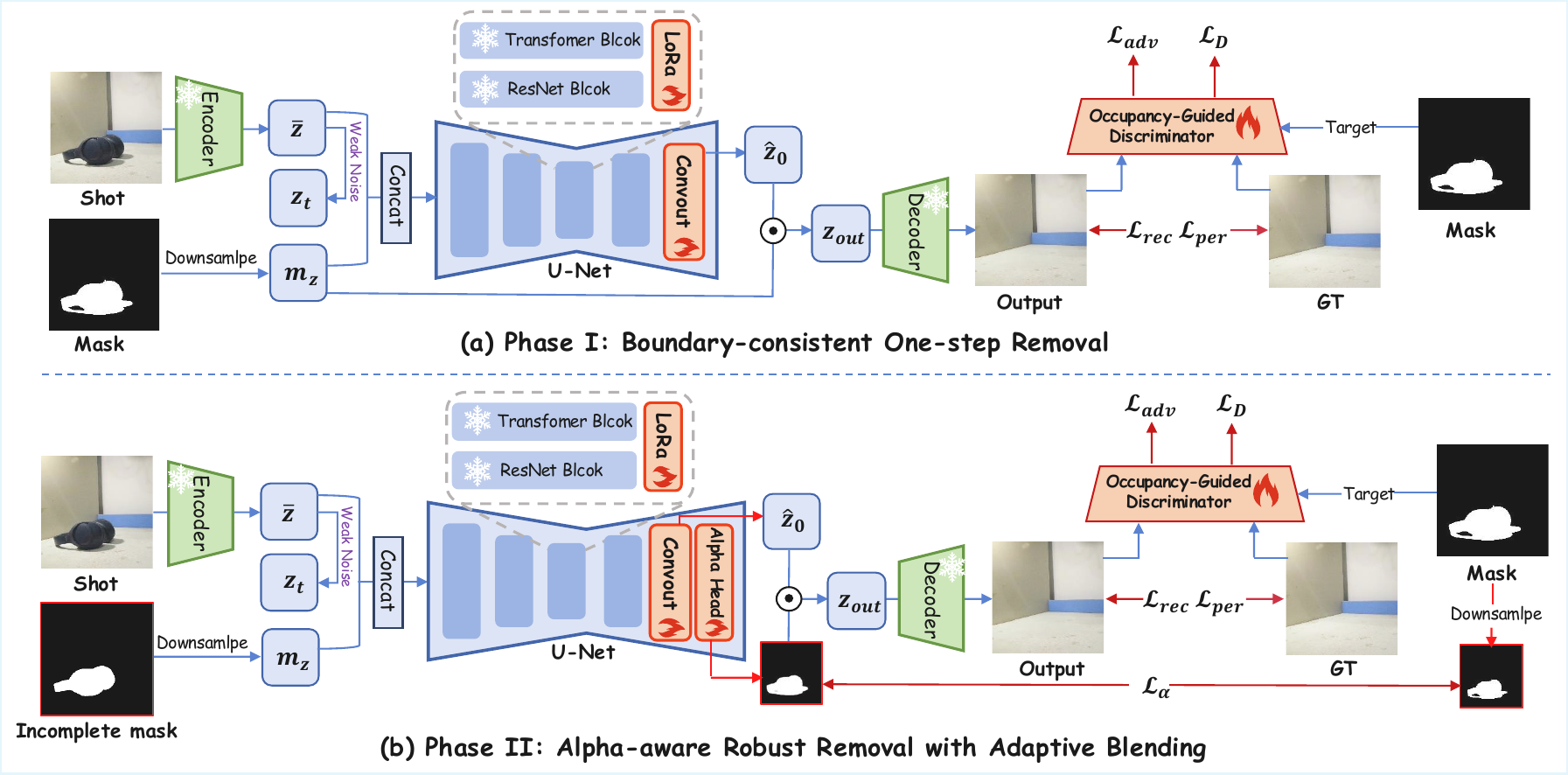}
    \caption{\textbf{Two-phase training curriculum.} Phase~I adapts a diffusion inpainting backbone with hard latent blending and occupancy-guided patch supervision for boundary-consistent one-step removal. Phase~II predicts a soft alpha map under incomplete-mask conditioning and performs adaptive blending to remove residual shadows and reflections beyond the provided mask.}
    \label{fig:train}
\end{figure}

OSOR performs one-step object-and-effect removal by restoring a clean background from an intermediate noised latent.
Training relies on effect-aware supervision and paired backgrounds, which we obtain with SAVP, a semantic-anchored verification pipeline over noisy instruction-based triplets, yielding the CORNE dataset with effect-aware masks.
We train OSOR in two phases, as summarized in Fig.~\ref{fig:train}.
Phase~I focuses on boundary-consistent one-step restoration under well-localized masks using an occupancy-guided multi-scale discriminator with patch-level targets.
Phase~II adds a lightweight alpha head and incomplete-mask conditioning to improve robustness to conservative or misaligned user masks.
Implementation details and verification thresholds are provided in the supplementary material.

\subsection{SAVP and the CORNE Dataset}
\label{sec:corne_dataset}

Effect-aware object removal requires paired backgrounds and masks that cover both the object and its visual effects such as cast shadows and reflections.
We introduce SAVP, a semantic-anchored verification pipeline that extracts reliable removal supervision from noisy instruction-based triplets.
We apply SAVP to the single-edit subset of NHR-Edit~\cite{kuprashevich2025nohumansrequired} with \emph{add} or \emph{remove} instructions.
For \emph{add}, we set $(I_{\text{shot}}, I_{\text{gt}})=(I_{\text{edit}}, I_{\text{orig}})$, and for \emph{remove} we swap the order.
SAVP verifies that the resulting pair exhibits localized and semantically consistent differences (Fig.~\ref{fig:dataset}), and outputs paired backgrounds with effect-aware masks to form CORNE.

\textbf{Semantic-anchored verification.}
Given a single-edit instruction triplet, SAVP forms an ordered image pair $(I_{\text{shot}}, I_{\text{gt}})$ and verifies that the visual difference is localized and semantically aligned with the instruction (Fig.~\ref{fig:dataset}). We compute a multi-feature difference heatmap from log-luminance, chromaticity, and gradient magnitude, then binarize and clean it to obtain a difference mask $m_{\text{diff}}$. Connected components in $m_{\text{diff}}$ yield candidate boxes $b_{\text{diff}}$. We run GroundingDINO~\cite{liu2023groundingdino} with the instruction text on $I_{\text{shot}}$ to obtain semantic boxes $b_{\text{sem}}$. We first apply a global rejection based on the fragmentation of dominant components and the noise ratio of small components to discard pairs with dispersed artifacts. We then traverse $b_{\text{diff}}$ in descending area and match each candidate to its best-overlapping semantic box in $b_{\text{sem}}$. A candidate is accepted as $b_{\text{val}}$ only if its best IoU exceeds a threshold and its area remains within a scale ratio bound, producing a refined mask $m_{\text{diff}}^{\text{val}}$. If a candidate violates the scale bound we discard the entire pair as a collapse case, otherwise we drop the candidate and continue. We keep a triplet only if at least one validated region exists. Implementation details and thresholds are deferred to the supplementary.

\textbf{Effect-aware mask synthesis.}
The validated difference mask $m_{\text{diff}}^{\text{val}}$ localizes the edit but can be fragmented and may miss parts of the object. We therefore obtain an object-core mask $m_{\text{obj}}$ with SAM2~\cite{ravi2024sam2} on $I_{\text{shot}}$ using $b_{\text{val}}$ as box prompts, and fuse it with the validated difference region (Fig.~\ref{fig:dataset}),
\begin{equation}
m_{\text{fuse}} = m_{\text{obj}} \cup m_{\text{diff}}^{\text{val}}.
\label{eq:mask_fusion}
\end{equation}
We optionally apply a lightweight dilation to $m_{\text{fuse}}$ to obtain the final effect-aware target mask $m_{\text{gt}}$ used in training.

\textbf{Effect decomposition for incomplete-mask conditioning.}
Phase~II requires tight object-core conditioning masks to simulate conservative user inputs. We define the effect residual on the pre-expansion fused mask,
\begin{equation}
m_{\text{eff}} = m_{\text{fuse}} \setminus m_{\text{obj}}.
\label{eq:effect_residual}
\end{equation}
We select effect-heavy cases by an effects ratio $\|m_{\text{eff}}\|_1 / \|m_{\text{fuse}}\|_1$. For effect-heavy cases, Phase~II samples the conditioning mask $m_{\text{in}}$ from a set of conservative masks that includes the tight object-core mask $m_{\text{obj}}$, while supervising with the effect-aware target $m_{\text{gt}}$.

\subsection{One-step Latent Restoration}
\label{sec:preliminaries}

OSOR builds on diffusion-family inpainting backbones (SDXL-Inpainting~\cite{podell2023sdxl} and FLUX Fill~\cite{blackforestlabs2024fluxfill}) and operates in the latent space of a pretrained VAE with encoder $E(\cdot)$ and decoder $D(\cdot)$.
Given an input \emph{shot} image $x$ and a user-provided mask $m$, we encode $\bar z=E(x)$ and apply forward noising at an intermediate noise level $t$~\cite{meng2021sdedit},
\begin{equation}
z_t = \alpha_t \bar z + \sigma_t \epsilon, \qquad \epsilon \sim \mathcal{N}(0,I),
\label{eq:forward_noising}
\end{equation}
where $\alpha_t$ and $\sigma_t$ are schedule coefficients.
The backbone is conditioned on the full input latent, the mask, and a fixed text embedding.
We use the constant prompt \emph{``Remove the instance of object''}, denote its embedding by $e$, and form $c=\langle \bar z, m, e\rangle$.
This exposes full scene context through $\bar z$ while using $m$ only to localize the intended edit.

Given $(z_t,c,t)$, the backbone predicts its native denoising output
\begin{equation}
u_\theta = f_\theta(z_t, c, t),
\label{eq:model_pred}
\end{equation}
and we obtain a one-step estimate of the clean latent via the corresponding one-step mapping,
\begin{equation}
\hat z_0 \;=\; \frac{z_t - \sigma_t\, u_\theta}{\alpha_t}.
\label{eq:onestep_recover}
\end{equation}

\subsection{Phase I: Boundary-consistent One-step Removal}
\label{sec:phase1}

Phase~I adapts a pretrained backbone for reliable one-step removal when the affected region is well localized.
We use CORNE supervision and take the effect-aware mask $m_{\text{gt}}$ as the target region.
In Phase~I we set the conditioning mask as $m=m_{\text{gt}}$.
Following Sec.~\ref{sec:preliminaries}, we obtain $z_t$ and the one-step prediction $\hat z_0$.
To preserve non-edited content exactly, we perform hard blending in latent space (Fig.~\ref{fig:train}a),
\begin{equation}
z_{\mathrm{out}} = m_z \odot \hat z_0 + (1-m_z)\odot \bar z,
\label{eq:phase1_hard_blend}
\end{equation}
where $\bar z = E(x)$ and $m_z$ is the backbone-specific mask representation used for latent blending.
We decode $\hat x = D(z_{\mathrm{out}})$.
Hard blending keeps the unmasked region identical to the input and restricts gradients from training objectives to the edited region, which stabilizes one-step adaptation.

\begin{figure}[t]
    \centering
    \includegraphics[width=1\linewidth]{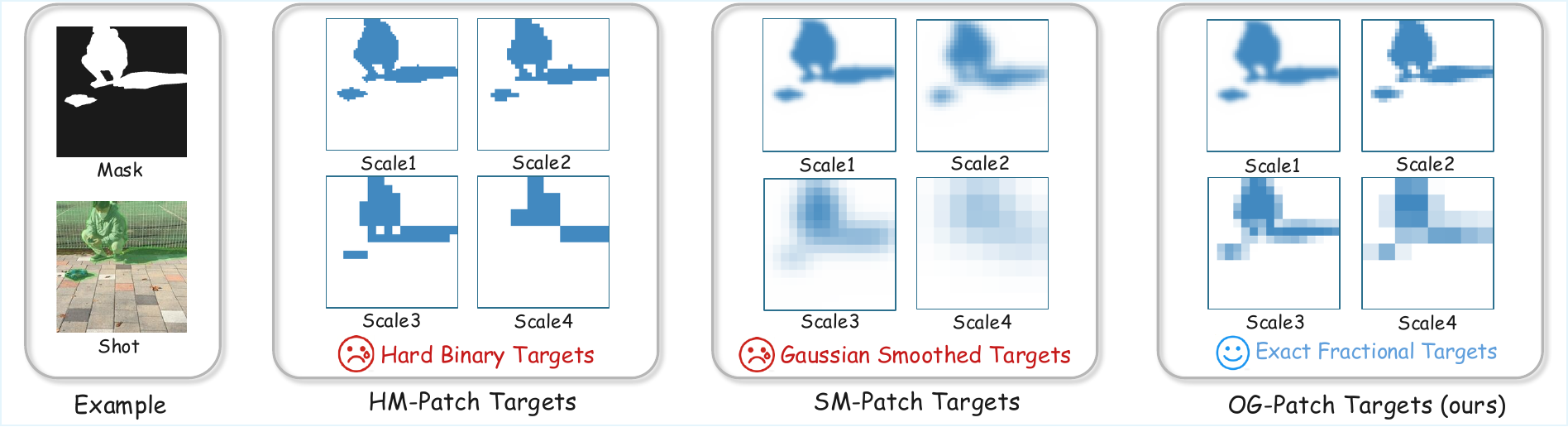}
    \caption{\textbf{Mask-derived patch targets for a four-scale discriminator.}
    Left shows the input mask and its overlay on the shot image for visualization.
    Right compares three target constructions at each scale.
    HM uses nearest-neighbor downsampling.
    SM applies Gaussian smoothing after downsampling.
    OG uses area pooling to produce fractional occupancies.
    Differences grow on coarser grids.}
    \label{fig:patchgan_viz}
\end{figure}

\textbf{Occupancy-guided multi-scale discriminator.}
Single-step restoration can preserve global structure but often shows seams near mask boundaries, where each discriminator patch mixes preserved context and synthesized content (Fig.~\ref{fig:patchgan_viz}).
We use a multi-scale patch discriminator in the PatchGAN family~\cite{isola2017pix2pix,suvorov2022lama,zeng2021aotgan}.
It consists of a frozen feature trunk $\phi$ and lightweight trainable patch heads $\{h_\xi^k\}_k$.
In our implementation, $\phi$ is a pretrained OpenCLIP ConvNeXt~\cite{radford2021clip,liu2022convnet,cherti2023scaling} that outputs multi-resolution features $f_k=\phi_k(x)$, and each head predicts a score map $D_\xi^k(x)=\sigma\!\big(h_\xi^k(f_k)\big)$.
We derive mask-based supervision targets at four scales, where target discretization becomes more pronounced at coarser heads (Fig.~\ref{fig:patchgan_viz}).

Mapping a binary mask to a coarse logit grid is ambiguous at boundary patches.
Nearest-neighbor downsampling produces hard labels and makes partially covered patches over-confident~\cite{suvorov2022lama}.
Gaussian smoothing yields soft targets but depends on the bandwidth $\sigma$, which is not tied to patch occupancy and becomes more consequential at larger downsampling factors~\cite{zeng2021aotgan}.
We instead compute an occupancy map $\tilde w_k\in[0,1]$ by area pooling the mask to each discriminator scale.
Each value equals the masked-area fraction within the patch of a logit location, giving exact fractional targets for boundary patches.

\textbf{Occupancy-guided objectives.}
We write the adversarial and reconstruction objectives using a spatial guidance map $w\in[0,1]^{H\times W}$.
In Phase~I, we set $w=m_{\text{gt}}$, and $\tilde w_k$ is obtained by area pooling $w$ to the $k$-th discriminator output resolution.
Let $x^{\mathrm{bg}}$ denote the paired ground-truth background.

The discriminator objective is
\begin{align}
\mathcal{L}_{D}(w)
&=
\sum_k \mathbb{E}\!\left[-\log D_\xi^k(x^{\mathrm{bg}})\right]
\nonumber\\
&\quad+
\sum_k \mathbb{E}\!\left[
-(1-\tilde w_k)\odot \log D_\xi^k(\hat x)
-\tilde w_k \odot \log\!\left(1-D_\xi^k(\hat x)\right)
\right]
+\lambda_{\mathrm{r1}}\mathcal{R}_{\mathrm{r1}}.
\label{eq:generic_disc}
\end{align}
We use an R1 regularizer on real samples~\cite{mescheder2018gan} and compute it on the head inputs since $\phi$ is frozen (see supplementary).

For the generator, we adopt the non-saturating form and normalize by the occupied area so the loss scale is insensitive to mask size,
\begin{equation}
\mathcal{L}_{\mathrm{adv}}(w)
=
\sum_k \mathbb{E}\!\left[
\frac{\sum_{p}\tilde w_k(p)\,\big(-\log D_\xi^k(\hat x)_p\big)}
{\sum_{p}\tilde w_k(p)+\varepsilon}
\right].
\label{eq:generic_adv}
\end{equation}
We also use a mask-normalized reconstruction term,
\begin{equation}
\mathcal{L}_{\mathrm{rec}}(w)
=
\mathbb{E}\left[
\frac{\left\| w \odot (\hat x - x^{\mathrm{bg}})\right\|_1}{\left\|w\right\|_1+\varepsilon}
\right],
\label{eq:generic_rec}
\end{equation}
together with $\mathcal{L}_{\mathrm{per}}=\mathrm{LPIPS}(\hat x, x^{\mathrm{bg}})$~\cite{zhang2018lpips}.
The generator objective is
\begin{equation}
\mathcal{L}_{G}(w)
=
\lambda_{\mathrm{rec}}\mathcal{L}_{\mathrm{rec}}(w)
+\lambda_{\mathrm{per}}\mathcal{L}_{\mathrm{per}}
+\lambda_{\mathrm{adv}}\mathcal{L}_{\mathrm{adv}}(w).
\label{eq:generic_gen}
\end{equation}

For parameter-efficient adaptation of large pretrained backbones, we update only lightweight adapters and the terminal output projection while keeping the remaining pretrained weights frozen.
Phase~I solves the adversarial game with $w=m_{\text{gt}}$ using
\begin{equation}
\min_{\theta}\ \max_{\xi}\
\mathcal{L}_{G}(m_{\text{gt}}) - \mathcal{L}_{D}(m_{\text{gt}}),
\label{eq:phase1_minmax}
\end{equation}
which is optimized by alternating updates of $\theta$ and $\xi$.

\subsection{Phase II: Alpha-aware Robust Removal with Adaptive Blending}
\label{sec:phase2}

Phase~I assumes the conditioning mask covers both the object and its effects.
In practice, user masks are often conservative or misaligned and frequently miss soft shadows and reflections.
Phase~II therefore trains OSOR with an \emph{incomplete} conditioning mask $m_{\text{in}}$ and predicts a soft alpha map for adaptive blending (Fig.~\ref{fig:train}b).

\textbf{Alpha prediction and adaptive blending.}
We augment the generator with a lightweight alpha head so that a single forward pass produces both the denoising prediction and an alpha logit map ~\cite{li2024drip,hu2024diffumatting,huang2024layerdiff}.
Concretely, we extend the terminal output projection of the backbone to emit an additional set of logits $\ell_\theta$ alongside its native denoising output.
For SDXL-Inpainting, we expand the final convolutional output layer of the U-Net; for FLUX Fill, we expand the final output projection of the transformer.
This design reuses all backbone computation and adds a small overhead.
Given $z_t$, $c$, and $t$, the generator predicts
\begin{equation}
(u_\theta,\; \ell_\theta)=f_\theta(z_t,c,t),\qquad \hat{\alpha}=\sigma(\ell_\theta),
\label{eq:phase2_alpha_pred}
\end{equation}
where $\hat{\alpha}\in[0,1]$ is predicted at the latent resolution and estimates the effective editing region.
For clarity, Fig.~\ref{fig:train}b visualizes $\hat{\alpha}$ as the alpha output of the head.
We recover $\hat z_0$ as in Sec.~\ref{sec:preliminaries} and replace hard blending with alpha compositing in latent space ~\cite{porter1984compositing,levin2008matting,xu2017deepmatting},
\begin{equation}
z_{\mathrm{out}}=\hat{\alpha}\odot \hat z_0 + (1-\hat{\alpha})\odot \bar z,
\label{eq:phase2_alpha_blend}
\end{equation}
with $\bar z=E(x)$.
Backbone-specific output parameterizations and the exact projection modifications are provided in the supplementary.

\begin{figure}[t]
    \centering
    \includegraphics[width=1.0\linewidth]{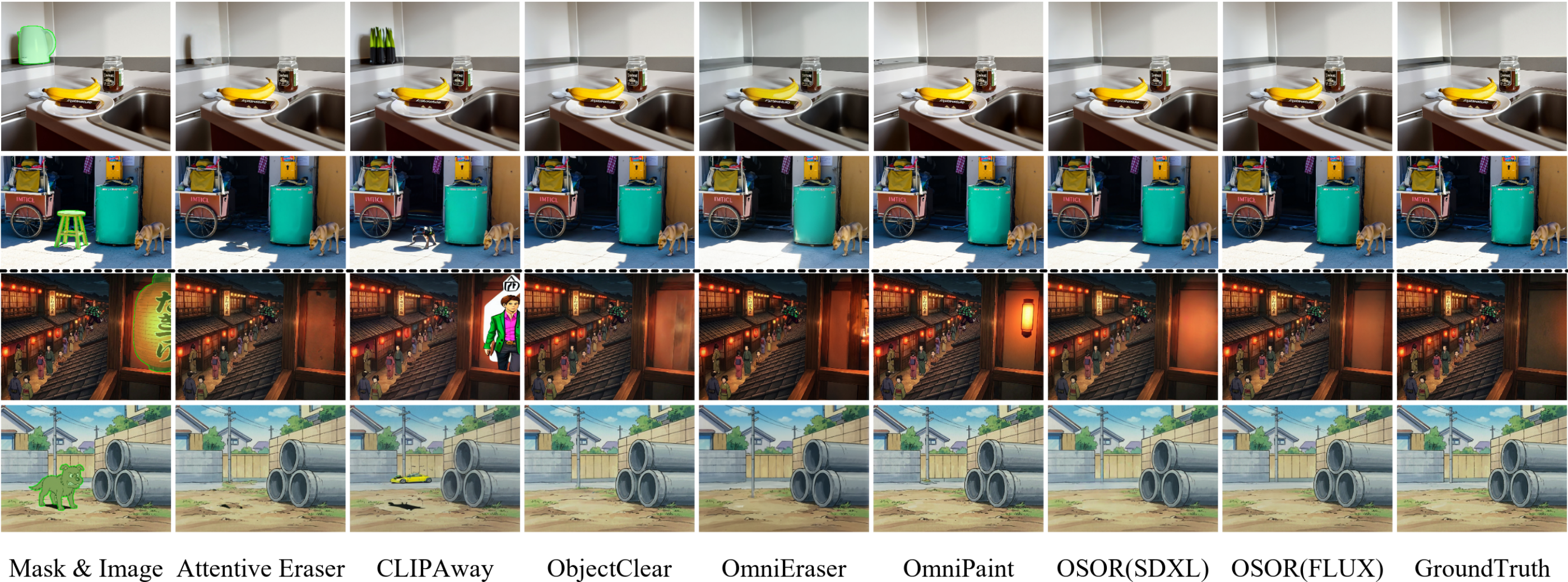} 
    \caption{Qualitative comparison of OSOR and existing methods on CORNE-Val and AnimeEraseBench.}
    \label{fig:corne_viz}
\end{figure}

\begin{wrapfigure}{r}{0.55\textwidth} \vspace{-15pt} \centering \includegraphics[width=1.0\linewidth]{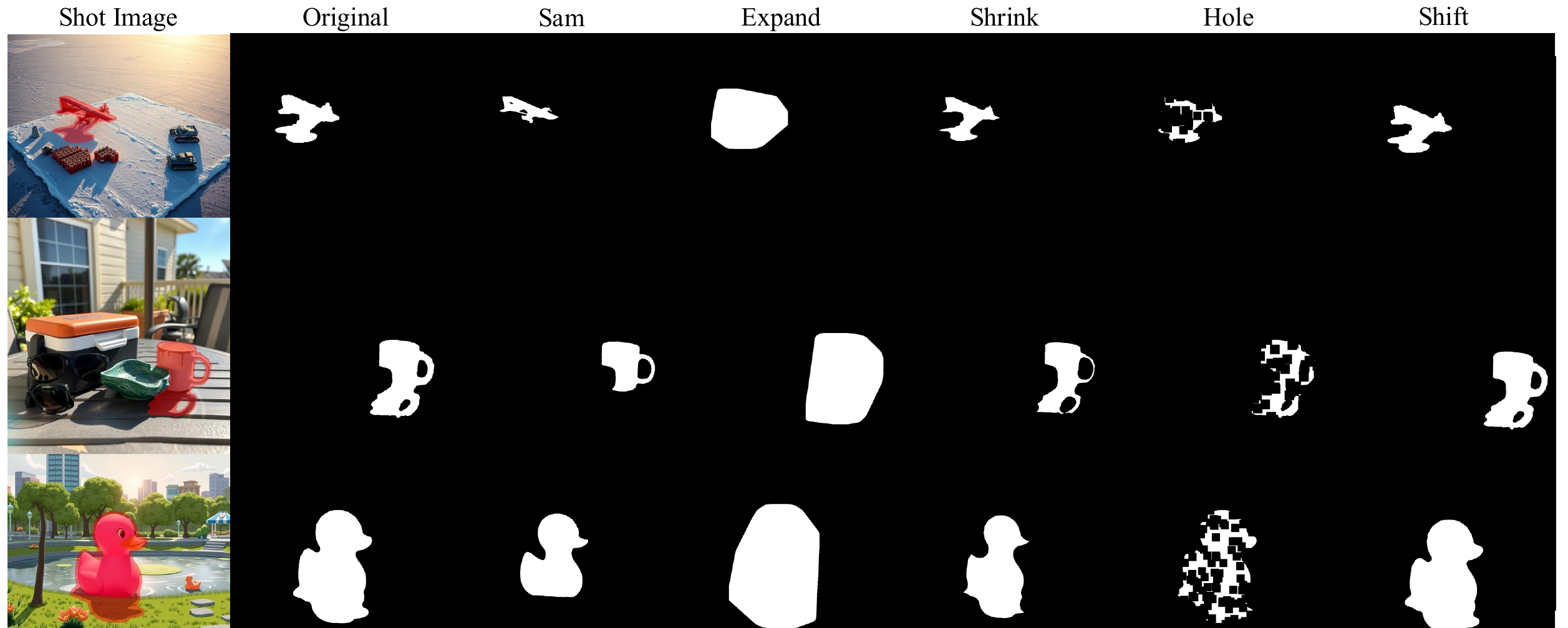} \caption{Examples of incomplete conditioning masks $m_{\text{in}}$ generated from object-core segmentation and simple geometric perturbations.} \label{fig:mask_viz} \vspace{-40pt} \end{wrapfigure}

\textbf{Incomplete-mask conditioning.}

In Phase~II, we replace the conditioning mask in $c$ by an incomplete mask $m_{\text{in}}$, using $c=\langle \bar z, m_{\text{in}}, e\rangle$.
We sample $m_{\text{in}}$ from a family of conservative masks derived from $m_{\text{gt}}$.
This family includes tight object-core masks $m_{\text{obj}}$ obtained via promptable segmentation, as well as simple perturbations of $m_{\text{gt}}$ such as dilation, erosion, translation, and random hole dropping, as illustrated in Fig.~\ref{fig:mask_viz}.
These conditioning masks intentionally under-cover the true affected region, so the model must infer missing effects from the image.

\begin{figure}[t]
    \centering
    \includegraphics[width=1.0\linewidth]{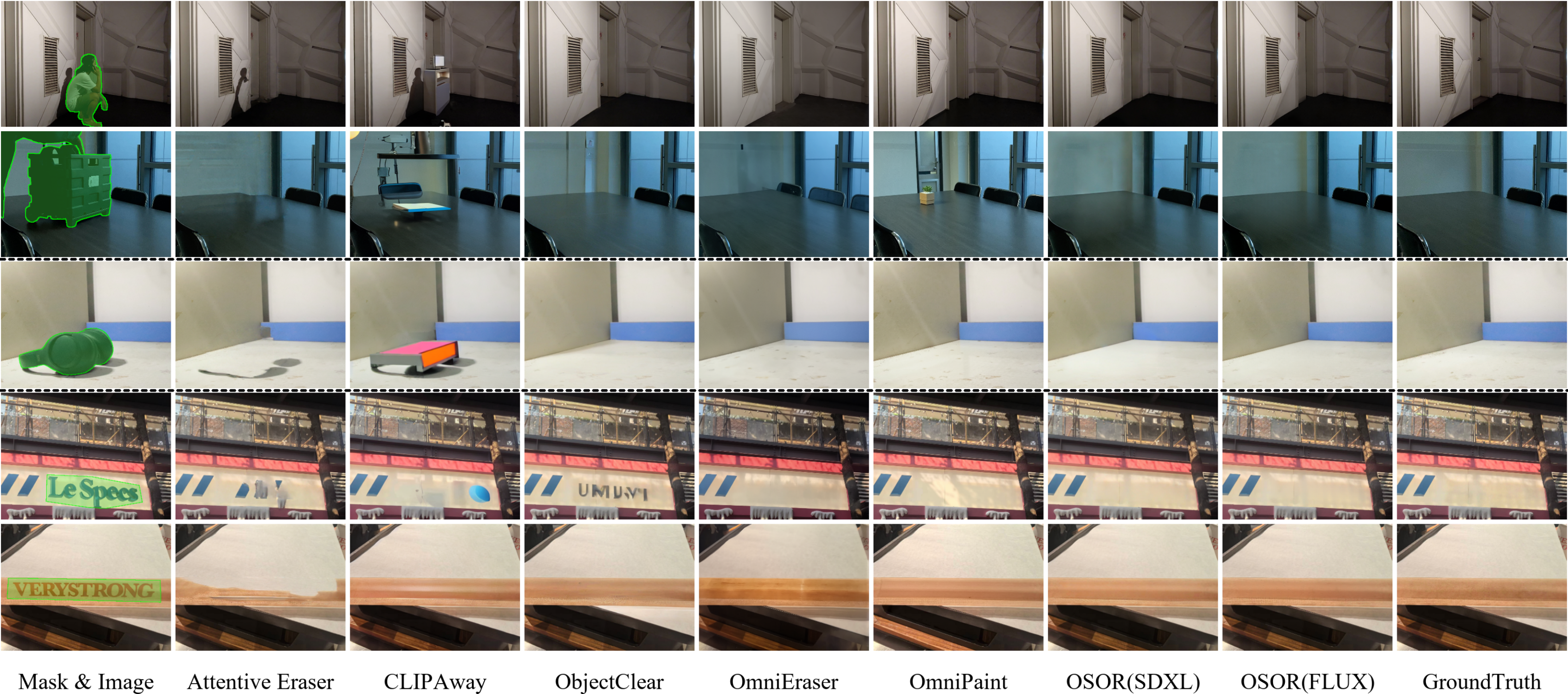} 
    \caption{Qualitative comparison of OSOR and existing methods on RORD-Val, RemovalBench and TextEraseBench.}
    \label{fig:rord_viz}
\end{figure}

\input{tab/table1}

\textbf{Alpha-guided training.}
A key design choice is that adversarial and reconstruction losses are always evaluated on the \emph{effect-aware target region} $m_{\text{gt}}$, rather than on the predicted $\hat{\alpha}$, to avoid degenerate solutions where the model reduces the loss by shrinking $\hat{\alpha}$.
Accordingly, we reuse the unified objectives in Eqs.~\eqref{eq:generic_disc}--\eqref{eq:generic_gen} with $w=m_{\text{gt}}$ in Phase~II as well.
The predicted alpha is used only in the latent compositing of Eq.~\eqref{eq:phase2_alpha_blend} and is explicitly supervised to match the effect-aware extent.
Let $m_{\text{gt}}^{z}$ denote $m_{\text{gt}}$ downsampled to the latent resolution (as illustrated in Fig.~\ref{fig:train}b). We add
\begin{equation}
\mathcal{L}_{\alpha}
=
\lambda_{\mathrm{bce}}\operatorname{BCE}(\ell_\theta,m_{\text{gt}}^{z})
+\lambda_{\mathrm{dice}}\operatorname{Dice}(\hat{\alpha},m_{\text{gt}}^{z}).
\label{eq:alpha_loss}
\end{equation}
This trains $\hat{\alpha}$ to recover the full effect-aware extent under incomplete conditioning, enabling removal of shadows and reflections beyond the user input.
The overall Phase~II objective is
\begin{equation}
\min_{\theta}\ \max_{\xi}\
\mathcal{L}_{G}(m_{\text{gt}}) + \mathcal{L}_{\alpha}
-\mathcal{L}_{D}(m_{\text{gt}}).
\label{eq:phase2_minmax}
\end{equation}
We initialize Phase~II from the Phase~I weights and maintain the same parameter-efficient adaptation strategy.
LoRA updates the main blocks, while the terminal projection that emits $(u_\theta,\ell_\theta)$ is fine-tuned to calibrate one-step output statistics.

%% file: tab/table1.tex
\begin{table*}[t]
\centering
\setlength{\tabcolsep}{3.8pt}
\renewcommand{\arraystretch}{1.12}

\caption{Quantitative comparison on paired-background benchmarks CORNE-Val, RORD-Val, and AnimeEraseBench under object-only masks and effect-aware masks.}
\label{tab:main}

\resizebox{\textwidth}{!}{%
\begin{tabular}{llc|cccccc|cccccc}
\toprule
\multirow{2}{*}{Dataset} & \multirow{2}{*}{Method} & \multirow{2}{*}{Latency (s)}
& \multicolumn{6}{c|}{Object Mask}
& \multicolumn{6}{c}{Object-Effect Mask} \\
\cmidrule(lr){4-9}\cmidrule(lr){10-15}
&  & & FID$\downarrow$ & CMMD$\downarrow$ & LPIPS$\downarrow$ & PSNR$\uparrow$ & SSIM$\uparrow$ & CFD$\downarrow$
   & FID$\downarrow$ & CMMD$\downarrow$ & LPIPS$\downarrow$ & PSNR$\uparrow$ & SSIM$\uparrow$ & CFD$\downarrow$ \\
\midrule

\multirow{9}{*}{\rotatebox{90}{CORNE-Val}} & OmniPaint
& 26.12 & \second{18.6077} & \second{0.0093} & \second{0.0563} & 28.436 & 0.9166 & 0.1950
& \second{19.111} & \second{0.0094} & \second{0.0603} & 27.950 & 0.9119 & 0.2806 \\
& ObjectClear
& 6.44 & 22.0457 & 0.0111 & 0.0956 & \second{29.207} & \second{0.9191} & 0.2040
& 22.3819 & 0.0113 & 0.0976 & \second{29.012} & \second{0.9166} & 0.3352 \\
& AttentiveEraser
& 8.69 & 25.4818 & 0.0193 & 0.0790 & 25.650 & 0.8636 & 0.2244
& 26.3795 & 0.0176 & 0.0825 & 25.602 & 0.8584 & 0.3867 \\
& FLUX-Fill
& 25.27 & 59.5678 & 0.0285 & 0.0824 & 23.124 & 0.9130 & 0.2766
& 53.2040 & 0.0260 & 0.0784 & 23.629 & 0.9119 & 0.7265 \\
& OmniEraser
& 20.53 & 60.3871 & 0.1079 & 0.1801 & 21.540 & 0.7623 & 0.2091
& 59.3292 & 0.1049 & 0.1800 & 21.720 & 0.7604 & 0.3250 \\
& SDXL-INP
& 6.01 & 75.1054 & 0.1506 & 0.2338 & 16.203 & 0.7149 & 0.2883
& 63.1940 & 0.1357 & 0.2276 & 16.804 & 0.7177 & 0.4665 \\
& CLIPAway
& 5.96 & 68.1043 & 0.1740 & 0.1864 & 21.446 & 0.7641 & 0.2746
& 51.2921 & 0.1587 & 0.1781 & 22.850 & 0.7676 & 0.4661 \\
& OSOR(SDXL)
& 0.42 & 29.3846 & 0.0249 & 0.1272 & 27.262 & 0.8371 & \second{0.1608}
& 29.2953 & 0.0255 & 0.1268 & 27.422 & 0.8376 & \best{0.2300} \\
& OSOR(FLUX)
& 0.80 & \best{12.3241} & \best{0.0073} & \best{0.0462} & \best{32.031} & \best{0.9373} & \best{0.1538}
& \best{12.5203} & \best{0.0069} & \best{0.0460} & \best{32.189} & \best{0.9377} & \second{0.2339} \\

\midrule

\multirow{9}{*}{\rotatebox{90}{RORD-Val}} & OmniPaint
& 16.51 & 27.5658 & \best{0.0668} & \best{0.1110} & 23.293 & 0.7899 & 0.3707
& 28.1457 & \best{0.0675} & \best{0.1147} & 23.078 & 0.7848 & 0.3754 \\
& ObjectClear
& 8.69 & \best{25.4667} & 0.1458 & 0.1206 & \best{25.804} & \best{0.8500} & 0.3248
& \best{26.3523} & 0.1456 & 0.1234 & \best{25.633} & \best{0.8462} & 0.3323 \\
& AttentiveEraser
& 8.68 & 36.5740 & 0.1951 & 0.1685 & 22.798 & 0.7023 & 0.3608
& 34.9941 & 0.1822 & 0.1688 & 23.080 & 0.6995 & 0.3558 \\
& FLUX-Fill
& 24.80 & 118.0815 & 0.5883 & 0.1821 & 20.171 & 0.8156 & 0.5415
& 111.9136 & 0.5451 & 0.1818 & 20.587 & 0.8127 & 0.5607 \\
& OmniEraser
& 25.99 & 43.2935 & 0.2409 & 0.2174 & 21.900 & 0.6550 & 0.3788
& 43.4599 & 0.2395 & 0.2183 & 21.881 & 0.6537 & 0.3788 \\
& SDXL-INP
& 6.38 & 95.2105 & 0.7893 & 0.3247 & 17.155 & 0.6466 & 0.5551
& 88.0085 & 0.7111 & 0.3234 & 17.398 & 0.6469 & 0.5788 \\
& CLIPAway
& 4.95 & 89.5176 & 0.9300 & 0.2962 & 19.447 & 0.5851 & 0.5805
& 81.5943 & 0.9031 & 0.2952 & 19.665 & 0.5850 & 0.5662 \\
& OSOR(SDXL)
& 0.42 & 35.5989 & 0.1854 & 0.1795 & 23.767 & 0.6827 & \second{0.2644}
& 35.6523 & 0.1858 & 0.1797 & 23.826 & 0.6823 & \second{0.2624} \\
& OSOR(FLUX)
& 0.62 & \second{27.3621} & \second{0.0952} & \second{0.1140} & \second{25.599} & \second{0.8175} & \best{0.2465}
& \second{28.1271} & \second{0.1016} & \second{0.1154} & \second{25.613} & \second{0.8163} & \best{0.2324} \\

\midrule

\multirow{9}{*}{\rotatebox{90}{AnimeEraseBench}} & OmniPaint
& 24.93 & \second{27.4034} & \second{0.0231} & \second{0.1129} & \second{25.542} & 0.8573 & 0.4369
& \second{26.8119} & \second{0.0223} & 0.1172 & 25.351 & 0.8515 & 0.4922 \\
& ObjectClear
& 10.18 & 36.6865 & 0.0241 & 0.1788 & 25.429 & 0.8409 & 0.3872
& 36.4259 & 0.0228 & 0.1800 & \second{25.530} & 0.8381 & 0.4406 \\
& AttentiveEraser
& 8.69 & 37.5448 & 0.0341 & 0.1333 & 22.587 & 0.7314 & 0.3802
& 37.7600 & 0.0337 & 0.1375 & 22.649 & 0.7259 & 0.4625 \\
& FLUX-Fill
& 25.19 & 76.5397 & 0.0558 & 0.1269 & 22.668 & \second{0.8597} & 0.4419
& 59.5985 & 0.0458 & \second{0.1115} & 23.960 & \second{0.8625} & 0.7075 \\
& OmniEraser
& 19.78 & 57.5116 & 0.1304 & 0.2339 & 19.190 & 0.6284 & 0.4249
& 56.6766 & 0.1220 & 0.2350 & 19.531 & 0.6232 & 0.4839 \\
& SDXL-INP
& 6.21 & 89.4030 & 0.1280 & 0.2754 & 14.827 & 0.6516 & 0.4952
& 70.5512 & 0.1081 & 0.2719 & 14.991 & 0.6516 & 0.5614 \\
& CLIPAway
& 5.96 & 88.9794 & 0.2142 & 0.3191 & 19.240 & 0.5748 & 0.4732
& 79.8083 & 0.2064 & 0.3143 & 19.581 & 0.5747 & 0.5293 \\
& OSOR(SDXL)
& 0.48 & 55.7614 & 0.0666 & 0.2367 & 23.090 & 0.6857 & \second{0.2820}
& 55.4380 & 0.0658 & 0.2368 & 23.130 & 0.6858 & \second{0.3713} \\
& OSOR(FLUX)
& 0.89 & \best{26.8352} & \best{0.0165} & \best{0.1070} & \best{27.780} & \best{0.8859} & \best{0.2703}
& \best{26.2539} & \best{0.0168} & \best{0.1077} & \best{27.871} & \best{0.8855} & \best{0.3505} \\

\bottomrule
\end{tabular}}
\end{table*}

%% file: sec/5_exp.tex
\section{Experiments}

We defer full implementation details, benchmark specifications, and metric definitions to the supplementary material.
Unless stated otherwise, all methods follow the same mask protocol and paired-background evaluation described there.
We compare with SDXL-Inpainting~\cite{podell2023sdxl} and FLUX Fill~\cite{blackforestlabs2024fluxfill} and with OmniEraser~\cite{wei2025omnieraser}, CLIPAway~\cite{ekin2024clipaway}, AttentiveEraser~\cite{sun2025attentiveeraser}, ObjectClear~\cite{zhao2025objectclear}, and OmniPaint~\cite{yu2025omnipaint} using official code and recommended settings, and we measure latency on an NVIDIA A100.

\begin{table}[t]
\begin{minipage}[t]{0.48\textwidth}
\centering
\caption{Noise-level ablation on RORD-Val with object-effect masks.}
\label{tab:timestep_ablation_rord_eff}
\setlength{\tabcolsep}{5pt}
\renewcommand{\arraystretch}{1.12}
\resizebox{\linewidth}{!}{
\begin{tabular}{c|cccccc}
\toprule
Timestep
& FID$\downarrow$
& CMMD$\downarrow$
& LPIPS$\downarrow$
& PSNR$\uparrow$
& SSIM$\uparrow$
& CFD$\downarrow$ \\
\midrule
200 & 60.1520 & 0.2463 & 0.1926 & 22.232 & 0.6718 & 0.3610 \\
400 & \textbf{50.0639} & \textbf{0.2143} & \textbf{0.1855} & \textbf{22.696} & \textbf{0.6741} & \textbf{0.3013} \\
600 & 55.4021 & 0.2263 & 0.1884 & 22.563 & 0.6720 & 0.3418 \\
800 & 51.8961 & 0.2505 & 0.1924 & 22.644 & 0.6710 & 0.3299 \\
\bottomrule
\end{tabular}
}
\end{minipage}
\hfill
\begin{minipage}[t]{0.48\textwidth}
\centering
\caption{Patch-target ablation on RORD-Val with object-effect masks.}
\label{tab:ablation_occ_patchgan}
\setlength{\tabcolsep}{3pt}
\resizebox{\linewidth}{!}{
\begin{tabular}{lccccc}
\toprule
Target & PSNR$\uparrow$ & SSIM$\uparrow$ & FID$\downarrow$ & LPIPS$\downarrow$ & CFD$\downarrow$ \\
\midrule
Hard Mask & 22.463 & 0.6739 & 56.1508 & 0.2007 & 0.3399 \\
Gaussian Soft & 21.992 & 0.6708 & 64.1290 & 0.2103 & 0.3430 \\
Occupancy & \textbf{22.696} & \textbf{0.6741} & \textbf{50.0639} & \textbf{0.1855} & \textbf{0.3013} \\
\bottomrule
\end{tabular}
}
\end{minipage}
\end{table}

\input{tab/table2}

\subsection{Comparison with Previous Methods}

\subsubsection{Quantitative Results.}
Table~\ref{tab:main} reports paired-background results under object-only masks and effect-aware masks. OSOR runs in under one second per image. On CORNE-Val, OSOR-FLUX achieves the best scores across all reported metrics under both mask settings, while OSOR-SDXL gives the lowest CFD under the effect-aware mask. On RORD-Val, OSOR-FLUX attains the lowest CFD under both masks and remains competitive on perceptual metrics. On AnimeEraseBench, OSOR-FLUX again ranks first across fidelity and perceptual measures under both masks. The results change little when switching between the two mask settings, which matches the goal of Phase~II training.
Table~\ref{tab:cross_domain} reports benchmarks evaluated under object-only masks. OSOR-FLUX performs best on TextEraseBench and OmniPaint-Bench for FID, CMMD, LPIPS, PSNR, and SSIM. On RemovalBench, OSOR-SDXL achieves the lowest CFD, and OSOR-FLUX remains competitive across the remaining metrics while retaining sub-second latency.

\subsubsection{Qualitative Results.}
Fig.~\ref{fig:rord_viz} and Fig.~\ref{fig:corne_viz} highlight common failure modes in existing methods.
CLIPAway often hallucinates new content inside the masked region.
AttentiveEraser removes the object but leaves cast shadows and reflections beyond an object-only mask.
OmniEraser, ObjectClear, and OmniPaint are generally strong on object removal and often suppress associated effects, yet occasional cases still exhibit mild residues, boundary inconsistencies, or unintended content.
OSOR more consistently removes both the object and its associated effects while preserving cleaner background structure and boundaries across the shown cases.

\subsection{Ablation Study}

\subsubsection{Noise level for one-step denoising.}
We study the noise level $t$ for one-step restoration on SDXL-Inpainting and evaluate on RORD-Val with object-effect masks.
We sweep $t\in\{200,400,600,800\}$ while keeping all other settings fixed, including the occupancy-guided multi-scale discriminator.
Table~\ref{tab:timestep_ablation_rord_eff} shows that $t=400$ yields the best overall trade-off across fidelity and perceptual metrics.
We use $t=400$ in all subsequent experiments unless stated otherwise.

\subsubsection{Occupancy-guided multi-scale discriminator.}
We ablate the construction of mask-derived patch targets on SDXL-Inpainting and evaluate on RORD-Val with object-effect masks.
We fix $t=400$ and compare hard targets from nearest-neighbor downsampling, Gaussian-smoothed targets, and our occupancy targets from area pooling.
Table~\ref{tab:ablation_occ_patchgan} shows that occupancy targets improve both perceptual and full-reference metrics, supporting fractional supervision at boundary patches.

\begin{figure}[t]
\begin{minipage}[c]{0.47\textwidth}
    \centering
    \includegraphics[width=1\linewidth]{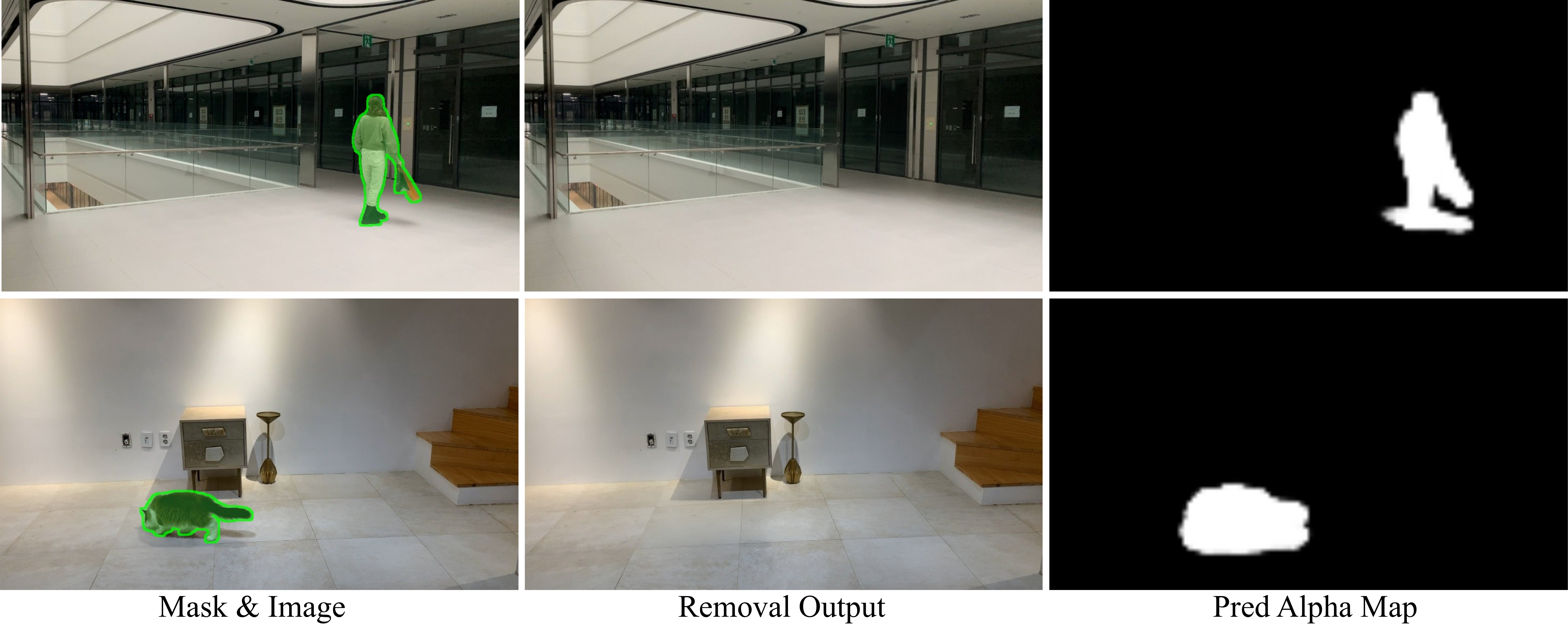}
    \caption{Alpha compositing under imperfect masks.}
    \label{fig:alpha_pred}
\end{minipage}
\hfill
\begin{minipage}[c]{0.5\textwidth}
\makeatletter\def\@captype{table}\makeatother
\centering
\caption{Ablation of alpha compositing on RORD-Val under object-only and effect-aware conditioning masks.}
\label{tab:ablation_alpha}
\resizebox{\linewidth}{!}{
\begin{tabular}{llcccccc}
\toprule
Conditioning mask & Method & PSNR$\uparrow$ & SSIM$\uparrow$ & FID$\downarrow$ & CMMD$\downarrow$ & LPIPS$\downarrow$ & CFD$\downarrow$ \\
\midrule
\multirow{2}{*}{Object-only $m_{\text{obj}}$} & Phase~I & 21.965 & 0.6726 & 59.8017 & 0.2764 & 0.2062 & 0.3133 \\
& Phase~II & \textbf{23.767} & \textbf{0.6827} & \textbf{35.5989} & \textbf{0.1854} & \textbf{0.1795} & \textbf{0.2644} \\
\midrule
\multirow{2}{*}{Effect-aware $m_{\text{gt}}$} & Phase~I & 22.696 & 0.6741 & 50.0639 & 0.2143 & 0.1855 & 0.3013 \\
& Phase~II & \textbf{23.826} & \textbf{0.6823} & \textbf{35.6523} & \textbf{0.1858} & \textbf{0.1797} & \textbf{0.2624} \\
\bottomrule
\end{tabular}
}
\end{minipage}
\end{figure}

\subsubsection{Alpha compositing with conservative masks.}
We evaluate whether the alpha head is needed for robust removal when the conditioning mask under-covers object effects.
We compare Phase~I with hard latent blending in Eq.~\eqref{eq:phase1_hard_blend} and Phase~II with alpha compositing in Eq.~\eqref{eq:phase2_alpha_blend} on RORD-Val under two conditioning masks.
The first uses the effect-aware mask $m_{\text{gt}}$.
The second uses an object-only mask $m_{\text{obj}}$ that excludes effect regions and serves as a conservative input. 
Table~\ref{tab:ablation_alpha} shows that Phase~II improves both fidelity and perceptual quality under both conditioning settings, with a clearer advantage when the conditioning mask is object-only.
Figure~\ref{fig:alpha_pred} visualizes the mechanism.
The predicted alpha extends into effect regions that are excluded from $m_{\text{obj}}$, enabling the model to modify a broader effective region than hard blending.

%% file: tab/table2.tex
\begin{table*}[t]
\centering
\setlength{\tabcolsep}{3.8pt}
\renewcommand{\arraystretch}{1.12}

\caption{Quantitative comparison under object-only masks on TextEraseBench, OmniPaint-Bench, and RemovalBench.}
\label{tab:cross_domain}

\resizebox{0.75\textwidth}{!}{%
\begin{tabular}{llc|cccccc}
\toprule
\multirow{2}{*}{Dataset} & \multirow{2}{*}{Method} & \multirow{2}{*}{Latency (s)}
& FID$\downarrow$ & CMMD$\downarrow$ & LPIPS$\downarrow$ & PSNR$\uparrow$ & SSIM$\uparrow$ & CFD$\downarrow$ \\
&  & & & & & & & \\
\midrule

\multirow{9}{*}{\rotatebox{90}{TextEraseBench}} & OmniPaint
& 24.89 & 29.3857 & 0.0348 & \second{0.1039} & 28.404 & 0.8811 & 0.2296 \\
& ObjectClear
& 10.16 & \second{29.1359} & \second{0.0293} & 0.1361 & \second{29.689} & \second{0.8945} & \best{0.1957} \\
& AttentiveEraser
& 8.69 & 34.6198 & 0.0397 & 0.1124 & 26.540 & 0.7974 & \second{0.1994} \\
& FLUX-Fill
& 25.27 & 85.1376 & 0.3464 & 0.1250 & 22.157 & 0.8781 & 0.4745 \\
& OmniEraser
& 19.80 & 53.7494 & 0.1129 & 0.2071 & 21.300 & 0.7134 & 0.2217 \\
& SDXL-INP
& 6.22 & 68.4836 & 0.1823 & 0.2383 & 19.300 & 0.7284 & 0.3422 \\
& CLIPAway
& 6.22 & 64.7926 & 0.3275 & 0.2440 & 23.106 & 0.7018 & 0.3011 \\
& OSOR(SDXL)
& 0.43 & 38.1242 & 0.0418 & 0.1800 & 27.748 & 0.7735 & 0.2128 \\
& OSOR(FLUX)
& 0.83 & \best{19.6169} & \best{0.0225} & \best{0.0803} & \best{31.997} & \best{0.9161} & 0.2066 \\

\midrule

\multirow{9}{*}{\rotatebox{90}{OmniPaint-Bench}} & OmniPaint
& 34.13 & \second{50.0921} & \second{0.0808} & \second{0.1522} & 23.455 & \second{0.8126} & 0.2700 \\
& ObjectClear
& 8.90 & 58.3634 & 0.1353 & 0.2357 & \second{23.543} & 0.7991 & 0.2535 \\
& AttentiveEraser
& 8.68 & 62.1880 & 0.1330 & 0.1732 & 22.406 & 0.7762 & 0.3290 \\
& FLUX-Fill
& 25.27 & 124.7098 & 0.3085 & 0.2056 & 19.460 & 0.7902 & 0.4750 \\
& OmniEraser
& 14.39 & 75.0069 & 0.1743 & 0.2502 & 22.063 & 0.7062 & 0.3194 \\
& SDXL-INP
& 6.16 & 115.0825 & 0.3088 & 0.2828 & 18.081 & 0.7059 & 0.5792 \\
& CLIPAway
& 5.96 & 117.2142 & 0.5298 & 0.3005 & 19.434 & 0.6972 & 0.5457 \\
& OSOR(SDXL)
& 0.42 & 67.4797 & 0.1967 & 0.2648 & 23.168 & 0.7419 & \best{0.2104} \\
& OSOR(FLUX)
& 1.06 & \best{49.1927} & \best{0.0703} & \best{0.1419} & \best{24.936} & \best{0.8330} & \second{0.2303} \\

\midrule

\multirow{9}{*}{\rotatebox{90}{RemovalBench}} & OmniPaint
& 30.59 & \best{41.1790} & \best{0.0273} & 0.2135 & 24.599 & 0.7488 & 0.2180 \\
& ObjectClear
& 10.16 & 49.2744 & 0.0335 & 0.2871 & \best{25.884} & \best{0.7726} & \second{0.1810} \\
& AttentiveEraser
& 8.69 & 80.3447 & 0.1788 & \best{0.1942} & 23.787 & 0.7002 & 0.2094 \\
& FLUX-Fill
& 25.27 & 178.6896 & 0.4454 & 0.2695 & 20.277 & 0.7099 & 0.3984 \\
& OmniEraser
& 14.39 & 65.4691 & 0.3066 & 0.2929 & 23.030 & 0.6712 & 0.1835 \\
& SDXL-INP
& 6.03 & 165.4925 & 0.4756 & 0.3388 & 16.983 & 0.6251 & 0.3227 \\
& CLIPAway
& 5.96 & 169.0450 & 0.4072 & 0.3787 & 20.461 & 0.6537 & 0.3569 \\
& OSOR(SDXL)
& 0.43 & 63.6405 & 0.0709 & 0.3313 & 24.676 & 0.6943 & \best{0.1585} \\
& OSOR(FLUX)
& 0.93 & \second{43.8899} & \second{0.0300} & \second{0.1972} & \second{25.803} & \second{0.7612} & \second{0.1810} \\

\bottomrule
\end{tabular}}
\end{table*}

%% file: sec/6_conclusion.tex
\section{Conclusion}
We presented OSOR, a one-step diffusion inpainting framework for effect-aware object removal. OSOR formulates removal as latent restoration from an intermediate noised latent and predicts the clean background in a single denoising pass. To improve boundary consistency in single-step training, we introduce an occupancy-guided multi-scale discriminator that uses fractional mask occupancies as patch-level targets. To handle conservative or misaligned user masks, we add a lightweight alpha head and train with incomplete-mask conditioning so the model can remove effects beyond the provided boundary. We further propose SAVP to extract effect-aware supervision from noisy instruction-based triplets and curate CORNE with 280K verified removal pairs, together with AnimeEraseBench and TextEraseBench for evaluation. Experiments show that OSOR reaches strong perceptual quality while reducing inference time by 4$\times$ to 30$\times$, and it processes 1024$\times$1024 images within one second on a single A100 GPU.

%% file: sec/7_supplementary.tex
\section{Supplementary Details of SAVP and CORNE}
\label{sec:supp_savp}

This section provides additional implementation details and dataset statistics for SAVP and CORNE.
We first describe semantic-anchored verification and effect-aware mask synthesis, then report aggregate statistics for CORNE and CORNE-Val, and finally present representative CORNE annotation cases.

\subsection{Algorithmic Details}
\label{sec:supp_savp_algo}

SAVP has two stages.
The first verifies that an instruction-based edit triplet yields a localized and semantically aligned removal pair.
The second synthesizes the effect-aware target mask and identifies the effect-heavy subset used in Phase~II.
Algorithm~\ref{alg:savp_verify} summarizes semantic-anchored verification.
Algorithm~\ref{alg:savp_mask} summarizes effect-aware mask synthesis and effect decomposition.

\begin{algorithm}[htbp]
\small
\caption{Semantic-anchored verification in SAVP}
\label{alg:savp_verify}
\KwIn{Single-edit instruction triplet $(I_{\text{orig}}, I_{\text{edit}}, p)$, where $p$ is the edit instruction}
\KwOut{Validated pair $(I_{\text{shot}}, I_{\text{gt}})$, validated boxes $b_{\text{val}}$, and refined difference mask $m_{\text{diff}}^{\text{val}}$; otherwise reject}

Determine the ordered pair $(I_{\text{shot}}, I_{\text{gt}})$ from the instruction type\;
\eIf{$p$ is an \emph{add} instruction}{
    $(I_{\text{shot}}, I_{\text{gt}}) \leftarrow (I_{\text{edit}}, I_{\text{orig}})$\;
}{
    \If{$p$ is a \emph{remove} instruction}{
        $(I_{\text{shot}}, I_{\text{gt}}) \leftarrow (I_{\text{orig}}, I_{\text{edit}})$\;
    }
    \Else{reject\;}
}

Compute normalized feature differences: log-luminance $\Delta L_{\log}$, chromaticity $\Delta ab$, and gradient-magnitude difference $\Delta \mathrm{Tex}$\;
Form the difference heatmap
\[
H = w_{L}\Delta L_{\log} + w_{ab}\Delta ab + w_{T}\Delta \mathrm{Tex}
\]
Threshold $H$ at $\tau_H$ to obtain a raw mask, then apply opening and closing with structuring radius $r_{\mathrm{morph}}$\;
Remove connected components with area smaller than $A_{\min}$, and fill holes with area at most $A_{\mathrm{hole}}$, yielding $m_{\text{diff}}$\;

Split connected components in $m_{\text{diff}}$ into dominant and noise components using the relative-area threshold $\alpha$\;
Reject the pair if the number of dominant components exceeds $N_{\max}$ or if the noise ratio exceeds $\tau_{\text{noise}}$\;

Convert dominant components into candidate boxes $b_{\text{diff}}$\;
Run GroundingDINO~\cite{liu2023groundingdino} on $I_{\text{shot}}$ with text query $p$, keeping up to $K_{\text{sem}}$ boxes with score at least $\tau_{\text{score}}$, to obtain semantic boxes $b_{\text{sem}}$\;
Initialize $b_{\text{val}} \leftarrow \varnothing$\;

Sort $b_{\text{diff}}$ by area in descending order\;
\ForEach{$b \in b_{\text{diff}}$}{
    Find the best-overlapping semantic box $b^\star \in b_{\text{sem}}$\;
    Compute
    \[
    v_{\text{iou}} = \mathrm{IoU}(b,b^\star), \qquad
    R = \frac{\mathrm{Area}(b)}{\mathrm{Area}(b^\star)}
    \]
    \uIf{$v_{\text{iou}} \ge \tau_{\text{iou}}$ and $R \le \tau_{\text{scale}}$}{
        accept $b$ and append it to $b_{\text{val}}$\;
    }
    \uElseIf{$R > \tau_{\text{scale}}$}{
        reject \tcp*[r]{collapse case}
    }
}

\If{$b_{\text{val}} = \varnothing$}{
    reject\;
}

Retain only connected components whose support from validated boxes exceeds $\tau_{\text{keep}}$ to construct $m_{\text{diff}}^{\text{val}}$\;
Return $(I_{\text{shot}}, I_{\text{gt}})$, $b_{\text{val}}$, and $m_{\text{diff}}^{\text{val}}$\;
\end{algorithm}

\begin{algorithm}[t]
\small
\caption{Effect-aware mask synthesis and effect decomposition}
\label{alg:savp_mask}
\KwIn{Validated pair $(I_{\text{shot}}, I_{\text{gt}})$, validated boxes $b_{\text{val}}$, refined difference mask $m_{\text{diff}}^{\text{val}}$}
\KwOut{Object-core mask $m_{\text{obj}}$, fused mask $m_{\text{fuse}}$, effect-aware target mask $m_{\text{gt}}$, effect residual $m_{\text{eff}}$, and a Phase~II candidate flag}

Run SAM2~\cite{ravi2024sam2} on $I_{\text{shot}}$ using $b_{\text{val}}$ as box prompts to obtain object-core proposals\;
Union all returned masks to form the object-core mask $m_{\text{obj}}$\;
Fuse the validated difference region with the object-core mask,
\[
m_{\text{fuse}} = m_{\text{obj}} \cup m_{\text{diff}}^{\text{val}}
\]
Expand $m_{\text{fuse}}$ by distance-transform-based area growth with ratio $r_{\text{dilate}}$ to obtain the final effect-aware target mask $m_{\text{gt}}$\;

Define the effect residual on the pre-expansion fused mask,
\[
m_{\text{eff}} = m_{\text{fuse}} \setminus m_{\text{obj}}
\]
and compute the effects ratio
\[
r_{\text{eff}} = \frac{\lVert m_{\text{eff}} \rVert_1}{\lVert m_{\text{fuse}} \rVert_1}
\]

Mark the sample as effect-heavy if $r_{\text{eff}} \ge \tau_{\text{eff}}$\;
For effect-heavy cases, construct conservative conditioning masks that include $m_{\text{obj}}$ and simple perturbations derived from $m_{\text{gt}}$\;
Return $m_{\text{obj}}$, $m_{\text{fuse}}$, $m_{\text{gt}}$, $m_{\text{eff}}$, and the Phase~II candidate flag\;
\end{algorithm}

\subsection{Implementation Constants}
\label{sec:supp_savp_thresholds}

Table~\ref{tab:savp_constants} summarizes the implementation constants used in SAVP.

\subsection{Dataset Statistics}
\label{sec:supp_savp_stats}

Table~\ref{tab:corne_stats} reports aggregate statistics for CORNE and CORNE-Val.
For training-set aggregation, we exclude part2 shards 10, 24, 30, and 31.
These shards are reserved for held-out sampling.
We randomly sample 1{,}500 pairs from the reserved shards, process them with the same pipeline, and obtain CORNE-Val with 219 samples.

\begin{table*}[t]
\centering

\begin{minipage}[t]{0.3\textwidth}
\footnotesize
\centering
\caption{SAVP implementation constants.}
\label{tab:savp_constants}
\setlength{\tabcolsep}{3pt}
\renewcommand{\arraystretch}{1.0}
\resizebox{\linewidth}{!}{%
\begin{tabular}{ll|ll}
\toprule
Symbol & Value & Symbol & Value \\
\midrule
$w_{L}$ & 0.6   & $\tau_{\text{noise}}$ & 0.3 \\
$w_{ab}$ & 0.3  & $\tau_{\text{score}}$ & 0.3 \\
$w_{T}$ & 0.1   & $K_{\text{sem}}$ & 3 \\
$\tau_H$ & 0.07 & $\tau_{\text{iou}}$ & 0.3 \\
$r_{\mathrm{morph}}$ & 3 & $\tau_{\text{scale}}$ & 2.0 \\
$A_{\min}$ & 2000 & $\tau_{\text{keep}}$ & 0.5 \\
$A_{\mathrm{hole}}$ & 500 & $r_{\text{dilate}}$ & 1.2 \\
$\alpha$ & 0.3 & $\tau_{\text{eff}}$ & 0.25 \\
$N_{\max}$ & 4 & & \\
\bottomrule
\end{tabular}%
}
\end{minipage}
\hfill
\begin{minipage}[t]{0.50\textwidth}
\footnotesize
\centering
\caption{CORNE and CORNE-Val statistics.}
\label{tab:corne_stats}
\setlength{\tabcolsep}{3pt}
\renewcommand{\arraystretch}{1.0}
\resizebox{\linewidth}{!}{%
\begin{tabular}{lcc}
\toprule
Statistic & Count & Ratio \\
\midrule
\multicolumn{3}{l}{\textit{Source corpus}} \\
Source samples & 680{,}088 & -- \\
Add-object samples & 213{,}204 & 31.35\% \\
Remove-object samples & 214{,}725 & 31.57\% \\
Add/remove total & 427{,}929 & 62.92\% \\
\midrule
\multicolumn{3}{l}{\textit{Retained training set}} \\
CORNE & 287{,}012 & 42.20\% of source \\
Phase~II subset & 67{,}726 & 23.60\% of CORNE \\
\midrule
\multicolumn{3}{l}{\textit{Held-out evaluation split}} \\
Held-out processed samples & 1{,}500 & -- \\
CORNE-Val & 219 & 14.60\% of held-out \\
\bottomrule
\end{tabular}%
}
\end{minipage}

\end{table*}

\subsection{Representative CORNE Annotation Cases}
\label{sec:supp_savp_vis}

\begin{figure*}[t]
    \centering
    \includegraphics[width=1\linewidth]{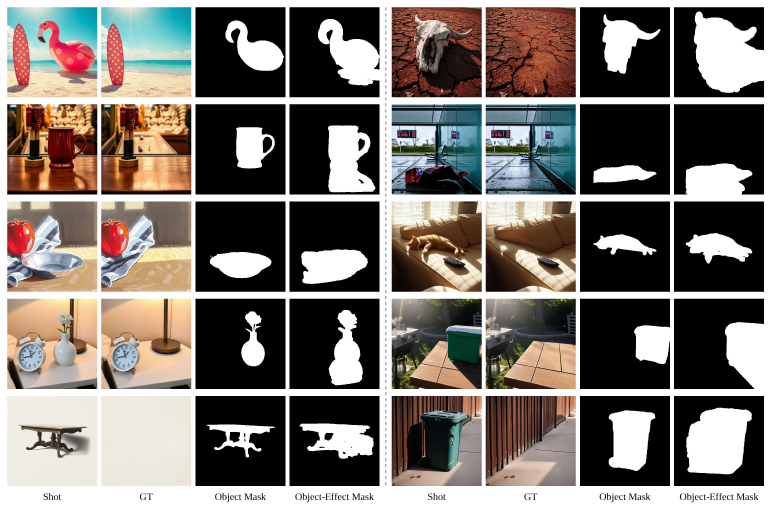}
    \caption{Representative CORNE annotation cases.
    Each row shows the input image $I_{\text{shot}}$, the paired background $I_{\text{gt}}$, the object-core mask $m_{\text{obj}}$, and the effect-aware target mask $m_{\text{gt}}$.
    The object-core mask provides tight object localization, while the effect-aware mask additionally covers object-induced visual effects such as cast shadows, reflections, and local residual traces.}
    \label{fig:corne_annotation_cases}
\end{figure*}

Figure~\ref{fig:corne_annotation_cases} shows representative annotation cases from CORNE.
Each sample contains the input image $I_{\text{shot}}$, the paired background $I_{\text{gt}}$, the tight object-core mask $m_{\text{obj}}$, and the effect-aware target mask $m_{\text{gt}}$.
Compared with $m_{\text{obj}}$, the effect-aware mask $m_{\text{gt}}$ additionally covers visual effects induced by the object, such as cast shadows, reflections, and local residual traces.
These examples illustrate the supervision structure used in Phase~I and the mask relationship underlying the incomplete-mask setting in Phase~II.

\section{Supplementary Training and Evaluation Details}
\label{sec:supp_exp}

\subsection{Implementation Details}
\label{sec:supp_impl}

We train OSOR on CORNE in two phases using two diffusion-family backbones,
SDXL-Inpainting~\cite{podell2023sdxl} and
FLUX Fill~\cite{blackforestlabs2024fluxfill}.

For SDXL-Inpainting, we resize each training image so that its shorter side is
512.
Phase~I uses LoRA~\cite{hu2022lora} with rank 256 and a global batch size of
16 on four NVIDIA A100 GPUs.
We train for 15K steps with a learning rate of $1\times10^{-5}$ for both the
generator and discriminator.
The loss weights are set to
$\lambda_{\mathrm{adv}}=0.3$,
$\lambda_{\mathrm{per}}=5$ using LPIPS~\cite{zhang2018lpips},
$\lambda_{\ell_1}=0.25$, and
$\lambda_{\mathrm{gp}}=60000$.
Phase~II keeps the same optimization settings and continues for 5K steps,
with additional alpha supervision using
$\lambda_{\mathrm{bce}}=1.0$ and
$\lambda_{\mathrm{dice}}=2.0$.
The complete two-phase training takes approximately 24 hours.

For FLUX Fill, inputs are resized to a multiple of 16.
Phase~I uses LoRA with rank 64 and a global batch size of 16 on eight NVIDIA
A100 GPUs.
We use the same learning rate and optimization schedule as for
SDXL-Inpainting, set $\lambda_{\ell_1}=0.5$ and $\lambda_{\mathrm{per}}=3$,
and keep the other loss weights unchanged.
Phase~II continues for 5K steps with the same alpha-supervision objectives.
The complete two-phase training takes approximately 30 hours.

Across both backbones, we update the LoRA modules and the terminal output
projections, including the alpha-output channels in Phase~II, while keeping
the remaining pretrained backbone parameters frozen.

\paragraph{Real-pair refinement.}
Starting from the Phase-II OSOR-FLUX checkpoint, we continue training for
1K steps on the \texttt{captured} subset of
OBER~\cite{zhao2025objectclear}.
All other settings are kept the same as in OSOR-FLUX Phase~II, including the
LoRA rank, trainable parameters, loss functions, loss weights, learning rate,
global batch size, and hardware configuration.
This additional training takes approximately 1.5 hours.

The refinement reduces the RORD-Val FID from 27.4/28.1 to 23.8/23.8 under
object-only/effect-aware masks.
It also reduces the FID on OmniPaint-Bench from 49.2 to 44.4 and that on
RemovalBench from 43.9 to 42.5.
Because the refinement only updates the model parameters, it does not change
the network architecture or the number of denoising steps at inference.

\subsection{Comparison Methods}
\label{sec:supp_baselines}

We compare OSOR with the baselines reported in the main paper.
We include the general diffusion inpainting backbones
SDXL-Inpainting~\cite{podell2023sdxl} and
FLUX Fill~\cite{blackforestlabs2024fluxfill}.
We also evaluate the object removal methods
OmniEraser~\cite{wei2025omnieraser},
CLIPAway~\cite{ekin2024clipaway},
Attentive Eraser~\cite{sun2025attentiveeraser},
ObjectClear~\cite{zhao2025objectclear}, and
OmniPaint~\cite{yu2025omnipaint}.
We use the official implementations and recommended settings whenever they
are available.

\paragraph{Training requirements.}
The compared methods follow different training protocols.
CLIPAway is training-free, while Attentive Eraser is tuning-free.
ObjectClear is trained for 100K iterations with a total batch size of 32 on
eight NVIDIA A100 GPUs; its wall-clock training time is not reported.
OmniEraser is trained for 130K steps with a batch size of 1 on a single
NVIDIA A800 GPU and reports a training time of approximately one day.
For OSOR, the complete two-phase training takes approximately 24 hours for
the SDXL-Inpainting backbone on four A100 GPUs and 30 hours for the FLUX Fill
backbone on eight A100 GPUs.
The additional 1K-step real-pair refinement takes approximately 1.5 hours.

These numbers describe the reported training requirements rather than a
compute-normalized comparison, since the methods differ in backbone, input
resolution, batch size, hardware, training data, and trainable parameters.

\paragraph{Inference speed.}
For runtime comparison, we measure the latency of all available methods on a
single NVIDIA A100 GPU using their official implementations and recommended
settings.
The real-pair refinement changes only the learned model parameters and
therefore does not affect the one-step inference procedure of OSOR-FLUX.

\subsection{Evaluation Benchmarks}
\label{sec:supp_benchmarks}

We evaluate on paired-background benchmarks under two mask settings at inference, an object mask $m_{\text{obj}}$ and an effect-aware mask $m_{\text{gt}}$ that additionally covers footprints such as shadows and reflections.
RORD-Val is built from RORD~\cite{sagong2022rord} by keeping one image per scene, yielding 343 samples, and re-annotating object and effect masks.
CORNE-Val contains 219 held-out CORNE samples with both mask types.
We further introduce \emph{AnimeEraseBench} with 157 samples and \emph{TextEraseBench} with 185 samples, covering stylized scenes and text removal with paired backgrounds and object and effect masks.
We also report results on OmniPaint-Bench~\cite{yu2025omnipaint} and RemovalBench~\cite{wei2025omnieraser}.

\subsection{Evaluation Metrics}
\label{sec:supp_metrics}

We report PSNR and SSIM as reference-based fidelity measures on paired-background benchmarks.
For perceptual quality, we use FID~\cite{heusel2017ttur}, CMMD~\cite{jayasumana2024rethinkingfid}, and LPIPS~\cite{zhang2018lpips}.
We additionally report CFD following OmniPaint~\cite{yu2025omnipaint}.

\subsection{Backbone-specific Alpha Head Implementation}
\label{sec:supp_alpha_head}

For both backbones, namely SDXL-Inpainting~\cite{podell2023sdxl} and FLUX Fill~\cite{blackforestlabs2024fluxfill}, the generator predicts $(u_\theta,\ell_\theta)=f_\theta(z_t,c,t)$ and the alpha map is obtained as $\hat{\alpha}=\sigma(\ell_\theta)$.
This subsection specifies how the alpha logits $\ell_\theta$ are parameterized for each backbone.

\noindent\textbf{SDXL-Inpainting.}
For SDXL-Inpainting, we modify the terminal U-Net output layer by expanding the final convolution from 4 to 5 output channels.
The first four channels retain the original denoising output and are initialized by copying the pretrained output convolution.
The additional fifth channel is newly initialized and serves as the alpha logit channel.
Let the modified U-Net output be
\[
y \in \mathbb{R}^{B\times 5\times H\times W}.
\]
We split it as
\[
u_\theta = y_{[:,\, :4,\, :, :]}, \qquad
\ell_\theta = y_{[:,\, 4:,\, :, :]}.
\]
The resulting $\ell_\theta$ is predicted directly at the latent resolution.
The modified terminal convolution is explicitly unfrozen and optimized jointly with the LoRA parameters.

\noindent\textbf{FLUX Fill.}
For FLUX Fill, we modify the terminal transformer output layer by expanding the final projection from 64 to 68 output dimensions.
The first 64 dimensions retain the original denoising output and are initialized by copying the pretrained output projection.
The additional 4 dimensions are newly initialized and represent alpha logits in the packed latent representation used by FLUX.
Let the modified transformer output be
\[
y^{\mathrm{pack}} \in \mathbb{R}^{N\times 68},
\]
where $N$ denotes the packed token dimension.
We split it as
\[
u_\theta^{\mathrm{pack}} = y^{\mathrm{pack}}_{[:,\, :64]}, \qquad
\ell_\theta^{\mathrm{pack}} = y^{\mathrm{pack}}_{[:,\, 64:]}.
\]
The packed alpha logits are then unpacked to the latent grid,
\[
\ell_\theta = \mathrm{Unpack}\!\left(\ell_\theta^{\mathrm{pack}}\right),
\qquad
\hat{\alpha}=\sigma(\ell_\theta),
\]
so that the final alpha map is also defined at the latent resolution.
The modified terminal projection is explicitly unfrozen and optimized jointly with the LoRA parameters.

\noindent\textbf{Parameter Overhead.}
This modification introduces only a negligible number of new parameters.
The added alpha outputs contribute 2,881 parameters for SDXL-Inpainting and 12,292 parameters for FLUX Fill.
In our implementation, the trainable terminal output layers are optimized in \texttt{float32}, so the corresponding parameter memory is approximately 11.3 KiB for SDXL-Inpainting and 48.0 KiB for FLUX Fill.
This overhead is negligible relative to the backbone size and does not introduce a meaningful memory burden in practice.

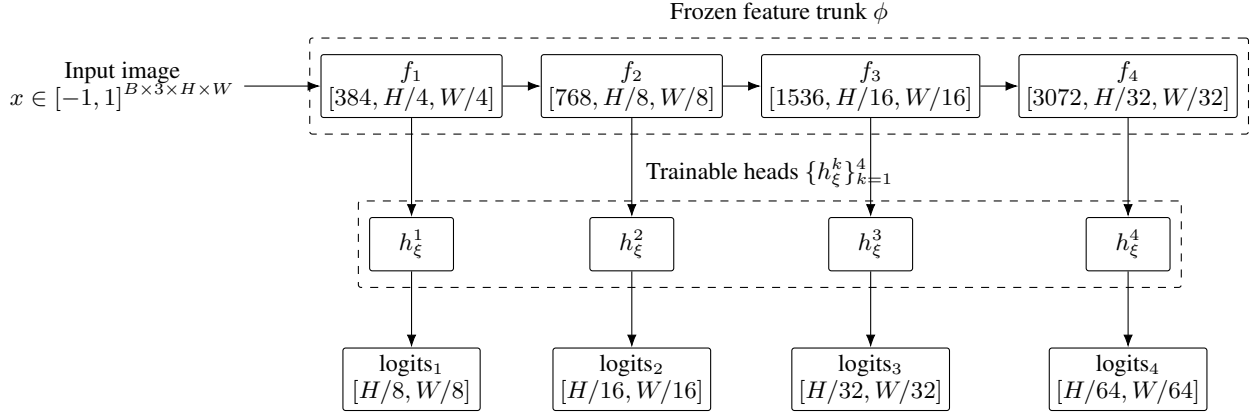
\begin{figure}[t]
\centering
\resizebox{\linewidth}{!}{%
\begin{tikzpicture}[
    >=Latex,
    font=\footnotesize,
    node distance=3mm and 6mm,
    box/.style={draw, rounded corners=1pt, align=center, inner sep=2.5pt, minimum height=8mm},
    stage/.style={box, minimum width=20mm},
    head/.style={box, minimum width=11mm, minimum height=7mm},
    logit/.style={box, minimum width=18mm},
    group/.style={draw, dashed, rounded corners=1pt, inner sep=4pt},
    lab/.style={align=center}
]

\node[lab] (input) {Input image\\$x\in[-1,1]^{B\times 3\times H\times W}$};

\node[stage, right=10mm of input] (s1) {$f_1$\\$[384,H/4,W/4]$};
\node[stage, right=5mm of s1] (s2) {$f_2$\\$[768,H/8,W/8]$};
\node[stage, right=5mm of s2] (s3) {$f_3$\\$[1536,H/16,W/16]$};
\node[stage, right=5mm of s3] (s4) {$f_4$\\$[3072,H/32,W/32]$};

\node[head, below=13mm of s1] (h1) {$h_\xi^1$};
\node[head, below=13mm of s2] (h2) {$h_\xi^2$};
\node[head, below=13mm of s3] (h3) {$h_\xi^3$};
\node[head, below=13mm of s4] (h4) {$h_\xi^4$};

\node[logit, below=10mm of h1] (l1) {logits$_1$\\$[H/8,W/8]$};
\node[logit, below=10mm of h2] (l2) {logits$_2$\\$[H/16,W/16]$};
\node[logit, below=10mm of h3] (l3) {logits$_3$\\$[H/32,W/32]$};
\node[logit, below=10mm of h4] (l4) {logits$_4$\\$[H/64,W/64]$};

\draw[->] (input.east) -- (s1.west);
\draw[->] (s1.east) -- (s2.west);
\draw[->] (s2.east) -- (s3.west);
\draw[->] (s3.east) -- (s4.west);

\draw[->] (s1.south) -- (h1.north);
\draw[->] (s2.south) -- (h2.north);
\draw[->] (s3.south) -- (h3.north);
\draw[->] (s4.south) -- (h4.north);

\draw[->] (h1.south) -- (l1.north);
\draw[->] (h2.south) -- (l2.north);
\draw[->] (h3.south) -- (l3.north);
\draw[->] (h4.south) -- (l4.north);

\node[group, fit=(s1)(s2)(s3)(s4), inner ysep=6pt] (g1) {};
\node[group, fit=(h1)(h2)(h3)(h4), inner ysep=6pt] (g2) {};

\node[font=\footnotesize, anchor=south] at ($(g1.north)+(0,1.5pt)$) {Frozen feature trunk $\phi$};
\node[font=\footnotesize, anchor=south] at ($(g2.north)+(0,1.5pt)$) {Trainable heads $\{h_\xi^k\}_{k=1}^4$};

\end{tikzpicture}%
}
\caption{Overall architecture of the occupancy-guided multi-scale discriminator.
A frozen feature trunk $\phi$ extracts four intermediate feature maps $f_k=\phi_k(x)$, which are processed by lightweight trainable heads $h_\xi^k$ to produce patch logits at four scales.}
\label{fig:supp_disc_arch}
\end{figure}

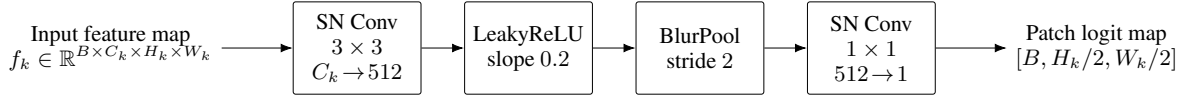
\begin{figure}[t]
\centering
\resizebox{0.95\linewidth}{!}{%
\begin{tikzpicture}[
    >=Latex,
    font=\footnotesize,
    node distance=3mm and 6mm,
    block/.style={draw, rounded corners=1pt, minimum width=18mm, minimum height=13mm, align=center},
    lab/.style={align=center}
]

\node[lab] (in) {Input feature map\\$f_k\in\mathbb{R}^{B\times C_k\times H_k\times W_k}$};

\node[block, right=10mm of in] (c1) {SN Conv\\$3\times3$\\$C_k\!\to\!512$};
\node[block, right=6mm of c1] (act) {LeakyReLU\\slope $0.2$};
\node[block, right=6mm of act] (blur) {BlurPool\\stride $2$};
\node[block, right=6mm of blur] (c2) {SN Conv\\$1\times1$\\$512\!\to\!1$};

\node[lab, right=10mm of c2] (out) {Patch logit map\\$[B,H_k/2,W_k/2]$};

\draw[->] (in.east) -- (c1.west);
\draw[->] (c1.east) -- (act.west);
\draw[->] (act.east) -- (blur.west);
\draw[->] (blur.east) -- (c2.west);
\draw[->] (c2.east) -- (out.west);

\end{tikzpicture}%
}
\caption{Structure of one trainable head $h_\xi^k$.
Each head applies spectral-normalized convolution, LeakyReLU, BlurPool downsampling, and a final $1\times1$ convolution to produce a single-channel patch logit map.}
\label{fig:supp_disc_head}
\end{figure}

\subsection{More Details of the Occupancy-guided Multi-scale Discriminator}
\label{sec:supp_discriminator}

Figure~\ref{fig:supp_disc_arch} summarizes the discriminator architecture.
It consists of a frozen feature trunk $\phi$ and lightweight trainable heads $\{h_\xi^k\}_{k=1}^4$.
Given an input image $x \in [-1,1]^{B\times 3\times H\times W}$, the image is first mapped to the CLIP image space, after which four intermediate feature maps
\[
f_k=\phi_k(x), \qquad k\in\{1,2,3,4\},
\]
are extracted.
In our implementation, $\phi$ is a pretrained OpenCLIP ConvNeXt visual encoder~\cite{cherti2023scaling,liu2022convnet}.
It produces four stages with channel dimensions $[384,768,1536,3072]$ and spatial resolutions $[H/4,W/4]$, $[H/8,W/8]$, $[H/16,W/16]$, and $[H/32,W/32]$, respectively.
Each stage is processed by one trainable head $h_\xi^k$, and the corresponding probability map is written as
\[
D_\xi^k(x)=\sigma\!\left(h_\xi^k(f_k)\right).
\]
In practice, the heads operate on logits, and the sigmoid is introduced only for notation.

Figure~\ref{fig:supp_disc_head} shows the structure of one trainable head.
All four heads share the same architecture and differ only in their input channel dimension.
Each head applies a spectral-normalized $3\times3$ convolution to project the incoming feature map to 512 channels, followed by a LeakyReLU activation with slope 0.2, a BlurPool layer with stride 2 for anti-aliased downsampling, and a spectral-normalized $1\times1$ convolution that produces a single-channel patch logit map.
After removing the singleton channel dimension, the four heads output logit maps at resolutions $[H/8,W/8]$, $[H/16,W/16]$, $[H/32,W/32]$, and $[H/64,W/64]$.

The feature trunk remains fixed throughout training, while only the multi-scale heads are updated.
This design reuses stable pretrained visual features and keeps the trainable part of the discriminator lightweight.
It also matches the objective design in the main text, where the R1 regularizer is evaluated on the head inputs because $\phi$ is frozen.

The discriminator architecture is shared across the HM, SM, and OG variants.
Their only difference lies in the construction of the supervision target $\tilde w_k$ at each scale.
HM uses nearest-neighbor downsampling, SM applies Gaussian smoothing after downsampling, and OG uses area pooling to produce exact fractional occupancies.
Therefore, the ablation in the main paper isolates the effect of target construction rather than changing the discriminator network itself.

\subsection{R1 Regularization on Head Inputs}
\label{sec:supp_r1}

The regularizer in Eq.~(17) is the R1 regularizer~\cite{mescheder2018gan} applied only to real samples.
Because the feature trunk $\phi$ is frozen, we evaluate R1 on the head inputs
\[
f_k=\phi_k(x^{\mathrm{bg}})
\]
rather than on the input image itself.
For each scale $k$, we compute the head logits $h_\xi^k(f_k)$ and differentiate the summed logits over all spatial positions with respect to $f_k$.
The resulting regularizer is
\begin{equation}
\mathcal{R}_{\mathrm{r1}}
=
\frac{1}{K}\sum_{k=1}^{K}
\mathbb{E}_{x^{\mathrm{bg}}}
\left[
\frac{1}{|f_k|}
\left\|
\nabla_{f_k}
\sum_{p} h_\xi^k(f_k)_p
\right\|_2^2
\right],
\label{eq:supp_r1}
\end{equation}
where $K$ is the number of discriminator scales, $p$ indexes spatial positions in the patch logit map, and $|f_k|$ is the number of elements in the feature tensor.

In implementation, the real feature maps are detached from the frozen trunk and treated as leaf tensors for gradient computation.
For each scale, we square and average the gradients over the full feature tensor, then average the result across scales.
This regularizes only the trainable heads while keeping the pretrained feature trunk fixed, which matches our discriminator parameterization and keeps the overhead low.

\begin{figure*}[t]
    \centering
    \includegraphics[width=1\linewidth]{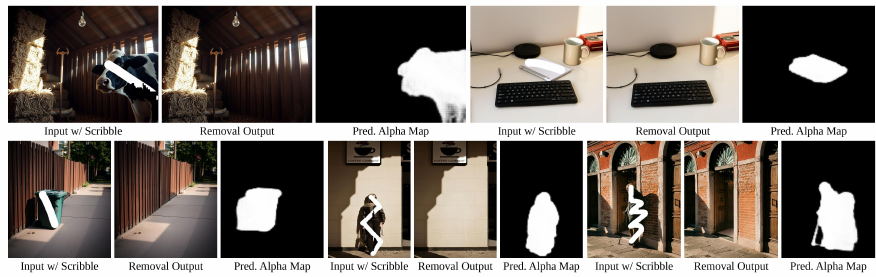}
    \caption{User scribble-guided removal examples.
For each case, we show the input image with user scribble, the removal result, and the predicted alpha map.
Starting from coarse user guidance, the model expands the effective removal region to cover the target object together with associated effects such as shadows and residual traces.}
    \label{fig:supp_user}
\end{figure*}

\section{Supplementary Details of EraseBench}
\label{sec:supp_erasebench}

\subsection{Dataset Construction of TextEraseBench}
\label{sec:supp_textbench}

The TextEraseBench dataset is constructed through a manual-to-automated pipeline designed for high-fidelity text removal.
We curate a diverse collection of real-world photographs and manually annotate target text regions with fine-grained bounding boxes.
These regions are then processed with Nano Banana 2 to remove the text while preserving background structure.
To ensure the quality of the paired backgrounds, each sample undergoes secondary verification to filter out artifacts and semantic inconsistencies.
The final benchmark contains 185 samples with paired backgrounds and both object and effect-aware masks.

\subsection{Dataset Construction of AnimeEraseBench}
\label{sec:supp_animebench}

AnimeEraseBench is developed through a synthesis-and-extraction pipeline tailored to stylized scenes.
We first generate diverse anime-style imagery and then remove selected foreground objects to obtain paired clean backgrounds.
We derive effect-aware masks from the differences between the source and background images, and use SAM2\cite{ravi2024sam2} together with manual box annotation to obtain object-core masks.
This dual-mask design enables evaluation under both object-only and effect-aware settings.
The final benchmark contains 157 samples.

\subsection{User Scribble-guided Removal}
\label{sec:supp_user}

Figure~\ref{fig:supp_user} shows user-guided removal examples under free-form scribble input.
For each sample, we show the input image overlaid with the user scribble, the removal result, and the predicted alpha map.
Although the scribble provides only coarse guidance, the model expands the removal region to cover the target object together with associated effects such as shadows and residual traces.

\subsection{Qualitative Examples}
\label{sec:supp_erasebench_vis}

\begin{figure*}[t]
    \centering
    \includegraphics[width=0.8\linewidth]{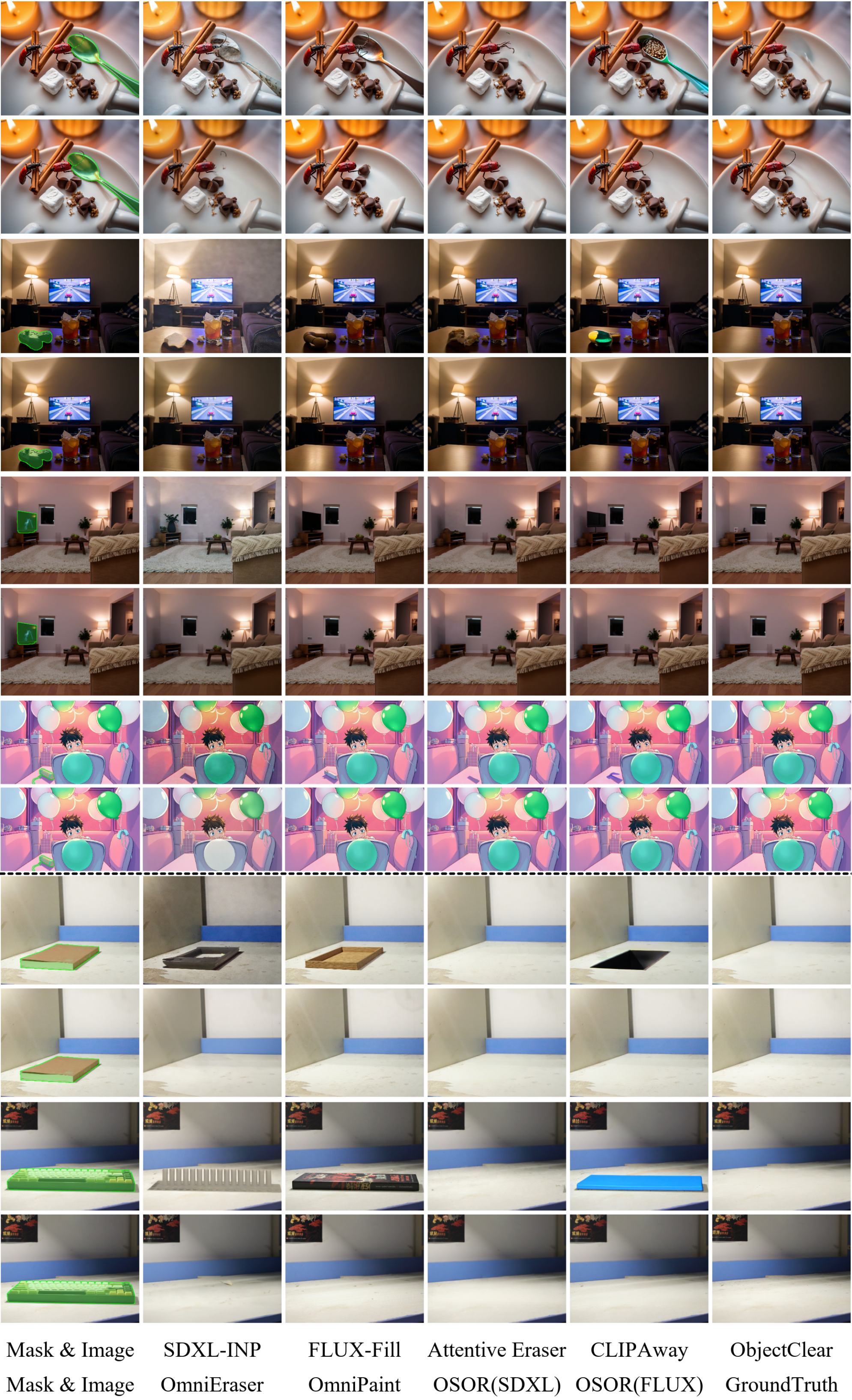}
    \caption{More qualitative comparisons of OSOR and existing methods on RemovalBench and CORNE-Val.}
    \label{fig:supp_main_qual_1}
\end{figure*}

\begin{figure*}[t]
    \centering
    \includegraphics[width=0.8\linewidth]{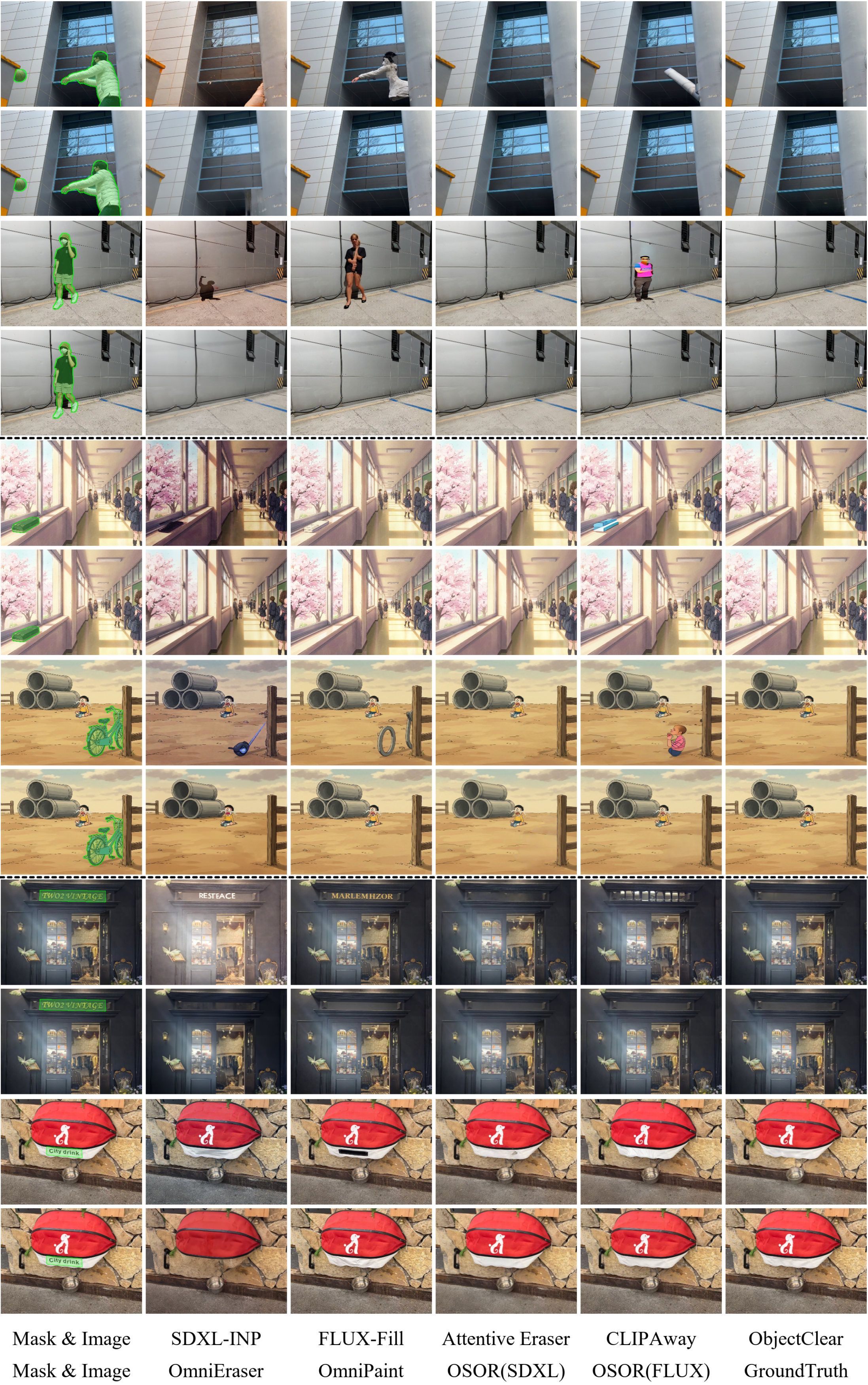}
    \caption{More qualitative comparisons of OSOR and existing methods on RORD-Val, AnimeEraseBench, and TextEraseBench.}
    \label{fig:supp_main_qual_2}
\end{figure*}

Figures~\ref{fig:supp_main_qual_1} and \ref{fig:supp_main_qual_2} present additional qualitative comparisons on representative samples.
These examples complement the main-paper visual results and cover diverse object categories, scene layouts, and effect types.

%% file: main.bib
@string(CVPR  = {CVPR})

@string(ICCV  = {ICCV})

@string(ECCV  = {ECCV})

@string(NeurIPS = {NeurIPS})

@string(ICML  = {ICML})

@string(ICLR  = {ICLR})

@string(BMVC  =	{BMVC})

@string(TOG   = {ACM TOG})

@inproceedings{goodfellow2014gan,
  author    = {Goodfellow, Ian J. and Pouget-Abadie, Jean and Mirza, Mehdi and Xu, Bing and Warde-Farley, David and Ozair, Sherjil and Courville, Aaron C. and Bengio, Yoshua},
  title     = {Generative Adversarial Nets},
  booktitle = NeurIPS,
  pages     = {2672--2680},
  year      = {2014}
}

@inproceedings{ho2020ddpm,
  author    = {Ho, Jonathan and Jain, Ajay and Abbeel, Pieter},
  title     = {Denoising Diffusion Probabilistic Models},
  booktitle = NeurIPS,
  pages     = {6840--6851},
  year      = {2020}
}

@inproceedings{pathak2016context,
  author    = {Pathak, Deepak and Kr{\"a}henb{\"u}hl, Philipp and Donahue, Jeff and Darrell, Trevor and Efros, Alexei A.},
  title     = {Context Encoders: Feature Learning by Inpainting},
  booktitle = CVPR,
  pages     = {2536--2544},
  year      = {2016}
}

@article{iizuka2017globally,
  author    = {Iizuka, Satoshi and Simo-Serra, Edgar and Ishikawa, Hiroshi},
  title     = {Globally and Locally Consistent Image Completion},
  journal   = TOG,
  volume    = {36},
  number    = {4},
  pages     = {107:1--107:14},
  year      = {2017}
}

@inproceedings{suvorov2022lama,
  author       = {Roman Suvorov and
                  Elizaveta Logacheva and
                  Anton Mashikhin and
                  Anastasia Remizova and
                  Arsenii Ashukha and
                  Aleksei Silvestrov and
                  Naejin Kong and
                  Harshith Goka and
                  Kiwoong Park and
                  Victor Lempitsky},
  title        = {Resolution-robust Large Mask Inpainting with Fourier Convolutions},
  booktitle    = {{IEEE/CVF} Winter Conference on Applications of Computer Vision, {WACV}
                  2022, Waikoloa, HI, USA, January 3-8, 2022},
  pages        = {3172--3182},
  year         = {2022},
  doi          = {10.1109/WACV51458.2022.00323}
}

@inproceedings{li2022mat,
  author    = {Li, Wenbo and Lin, Zhe and Zhou, Kun and Qi, Lu and Wang, Yi and Jia, Jiaya},
  title     = {MAT: Mask-Aware Transformer for Large Hole Image Inpainting},
  booktitle = CVPR,
  pages     = {10748--10758},
  year      = {2022}
}

@inproceedings{dong2022zits,
  author    = {Dong, Qiaole and Cao, Chenjie and Fu, Yanwei},
  title     = {Incremental Transformer Structure Enhanced Image Inpainting with Masking Positional Encoding},
  booktitle = CVPR,
  pages     = {11348--11358},
  year      = {2022}
}

@inproceedings{rombach2022ldm,
  title     = {High-Resolution Image Synthesis with Latent Diffusion Models},
  author    = {Rombach, Robin and Blattmann, Andreas and Lorenz, Dominik and Esser, Patrick and Ommer, Bj{\"o}rn},
  booktitle = CVPR,
  pages     = {10674--10685},
  year      = {2022}
}

@inproceedings{lugmayr2022repaint,
  title     = {RePaint: Inpainting using Denoising Diffusion Probabilistic Models},
  author    = {Lugmayr, Andreas and Danelljan, Martin and Romero, Andres and Yu, Fisher and Timofte, Radu and Van Gool, Luc},
  booktitle = CVPR,
  pages     = {11451--11461},
  year      = {2022}
}

@inproceedings{avrahami2022blended,
  title     = {Blended Diffusion for Text-Driven Editing of Natural Images},
  author    = {Avrahami, Omri and Lischinski, Dani and Fried, Ohad},
  booktitle = CVPR,
  pages     = {18187--18197},
  year      = {2022}
}

@inproceedings{brooks2022instructpix2pix,
  author    = {Brooks, Tim and Holynski, Aleksander and Efros, Alexei A.},
  title     = {InstructPix2Pix: Learning to Follow Image Editing Instructions},
  booktitle = CVPR,
  pages     = {18392--18402},
  year      = {2023}
}

@inproceedings{podell2023sdxl,
 author       = {Dustin Podell and
                  Zion English and
                  Kyle Lacey and
                  Andreas Blattmann and
                  Tim Dockhorn and
                  Jonas M{\"{u}}ller and
                  Joe Penna and
                  Robin Rombach},
  title        = {{SDXL:} Improving Latent Diffusion Models for High-Resolution Image
                  Synthesis},
  booktitle    = {The Twelfth International Conference on Learning Representations,
                  {ICLR} 2024, Vienna, Austria, May 7-11, 2024},
  year         = {2024},
}

@misc{blackforestlabs2024fluxfill,
  author = {{Black Forest Labs}},
  title={FLUX},
  year={2024},
  howpublished={\url{https://github.com/black-forest-labs/flux}},
}

@inproceedings{meng2021sdedit,
  author       = {Chenlin Meng and
                  Yutong He and
                  Yang Song and
                  Jiaming Song and
                  Jiajun Wu and
                  Jun{-}Yan Zhu and
                  Stefano Ermon},
  title        = {SDEdit: Guided Image Synthesis and Editing with Stochastic Differential
                  Equations},
  booktitle    = {The Tenth International Conference on Learning Representations, {ICLR}
                  2022, Virtual Event, April 25-29, 2022},
  year         = {2022}
}

@inproceedings{sagong2022rord,
   author       = {Min{-}Cheol Sagong and
                  Yoon{-}Jae Yeo and
                  Seung{-}Won Jung and
                  Sung{-}Jea Ko},
  title        = {{RORD:} {A} Real-world Object Removal Dataset},
  booktitle    = {33rd British Machine Vision Conference 2022, {BMVC} 2022, London,
                  UK, November 21-24, 2022},
  pages        = {542},
  year         = {2022},
}

@inproceedings{winter2024objectdrop,
  author       = {Daniel Winter and
                  Matan Cohen and
                  Shlomi Fruchter and
                  Yael Pritch and
                  Alex Rav{-}Acha and
                  Yedid Hoshen},
  title        = {ObjectDrop: Bootstrapping Counterfactuals for Photorealistic Object
                  Removal and Insertion},
  booktitle    = {Computer Vision - {ECCV} 2024 - 18th European Conference, Milan, Italy,
                  September 29-October 4, 2024, Proceedings, Part {LXXVII}},
  volume       = {15135},
  pages        = {112--129},
  year         = {2024}
}

@article{kuprashevich2025nohumansrequired,
  author       = {Maksim Kuprashevich and
                  Grigorii Alekseenko and
                  Irina Tolstykh and
                  Georgii Fedorov and
                  Bulat Suleimanov and
                  Vladimir Dokholyan and
                  Aleksandr Gordeev},
  title        = {NoHumansRequired: Autonomous High-Quality Image Editing Triplet Mining},
  journal      = {CoRR},
  volume       = {abs/2507.14119},
  year         = {2025},
  doi          = {10.48550/ARXIV.2507.14119},
  eprinttype    = {arXiv},
  eprint       = {2507.14119}
}

@inproceedings{jiang2025smarteraser,
  author       = {Longtao Jiang and
                  Zhendong Wang and
                  Jianmin Bao and
                  Wengang Zhou and
                  Dongdong Chen and
                  Lei Shi and
                  Dong Chen and
                  Houqiang Li},
  title        = {SmartEraser: Remove Anything from Images using Masked-Region Guidance},
  booktitle    = {{IEEE/CVF} Conference on Computer Vision and Pattern Recognition,
                  {CVPR} 2025, Nashville, TN, USA, June 11-15, 2025},
  pages        = {24452--24462},
  year         = {2025},
  doi          = {10.1109/CVPR52734.2025.02277},
}

@article{zhao2025objectclear,
  author       = {Jixin Zhao and
                  Shangchen Zhou and
                  Zhouxia Wang and
                  Peiqing Yang and
                  Chen Change Loy},
  title        = {ObjectClear: Complete Object Removal via Object-Effect Attention},
  journal      = {CoRR},
  volume       = {abs/2505.22636},
  year         = {2025},
  doi          = {10.48550/ARXIV.2505.22636},
}

@misc{wei2025omnieraser,
  title={OmniEraser: Remove Objects and Their Effects in Images with Paired Video-Frame Data},
  author={Runpu Wei and Zijin Yin and Shuo Zhang and Lanxiang Zhou and Xueyi Wang and Chao Ban and Tianwei Cao and Hao Sun and Zhongjiang He and Kongming Liang and Zhanyu Ma},
  year={2025},
  eprint={2501.07397},
  archivePrefix={arXiv},
  primaryClass={cs.CV},
  url={https://arxiv.org/abs/2501.07397},
}

@inproceedings{ekin2024clipaway,
  author       = {Yigit Ekin and
                  Ahmet Burak Yildirim and
                  Erdem Eren Caglar and
                  Aykut Erdem and
                  Erkut Erdem and
                  Aysegul Dundar},
  editor       = {Amir Globersons and
                  Lester Mackey and
                  Danielle Belgrave and
                  Angela Fan and
                  Ulrich Paquet and
                  Jakub M. Tomczak and
                  Cheng Zhang},
  title        = {CLIPAway: Harmonizing focused embeddings for removing objects via
                  diffusion models},
  booktitle    = {Advances in Neural Information Processing Systems 38: Annual Conference
                  on Neural Information Processing Systems 2024, NeurIPS 2024, Vancouver,
                  BC, Canada, December 10 - 15, 2024},
  year         = {2024}
}

@inproceedings{sun2025attentiveeraser,
  author       = {Wenhao Sun and
                  Xue{-}Mei Dong and
                  Benlei Cui and
                  Jingqun Tang},
  title        = {Attentive Eraser: Unleashing Diffusion Model's Object Removal Potential
                  via Self-Attention Redirection Guidance},
  booktitle    = {AAAI-25, Sponsored by the Association for the Advancement of Artificial
                  Intelligence, February 25 - March 4, 2025, Philadelphia, PA, {USA}},
  pages        = {20734--20742},
  year         = {2025},
  doi          = {10.1609/AAAI.V39I19.34285}
}

@inproceedings{liu2025erasediffusion,
  author       = {Yi Liu and
                  Hao Zhou and
                  Benlei Cui and
                  Wenxiang Shang and
                  Ran Lin},
  title        = {Erase Diffusion: Empowering Object Removal Through Calibrating Diffusion
                  Pathways},
  booktitle    = {{IEEE/CVF} Conference on Computer Vision and Pattern Recognition,
                  {CVPR} 2025, Nashville, TN, USA, June 11-15, 2025},
  pages        = {2418--2427},
  year         = {2025},
  doi          = {10.1109/CVPR52734.2025.00231}
}

@inproceedings{salimans2022progressive,
  author       = {Tim Salimans and
                  Jonathan Ho},
  title        = {Progressive Distillation for Fast Sampling of Diffusion Models},
  booktitle    = {The Tenth International Conference on Learning Representations, {ICLR}
                  2022, Virtual Event, April 25-29, 2022},
  year         = {2022},
  url          = {https://openreview.net/forum?id=TIdIXIpzhoI}
}

@inproceedings{song2023consistency,
  author       = {Yang Song and
                  Prafulla Dhariwal and
                  Mark Chen and
                  Ilya Sutskever},
  title        = {Consistency Models},
  booktitle    = {International Conference on Machine Learning, {ICML} 2023, 23-29 July
                  2023, Honolulu, Hawaii, {USA}},
  volume       = {202},
  pages        = {32211--32252},
  year         = {2023}
}

@article{luo2023lcm,
  author       = {Simian Luo and
                  Yiqin Tan and
                  Longbo Huang and
                  Jian Li and
                  Hang Zhao},
  title        = {Latent Consistency Models: Synthesizing High-Resolution Images with
                  Few-Step Inference},
  journal      = {CoRR},
  volume       = {abs/2310.04378},
  year         = {2023},
  doi          = {10.48550/ARXIV.2310.04378},
  eprinttype    = {arXiv},
  eprint       = {2310.04378}
}

@inproceedings{yin2024dmd,
  author       = {Tianwei Yin and
                  Micha{\"{e}}l Gharbi and
                  Richard Zhang and
                  Eli Shechtman and
                  Fr{\'{e}}do Durand and
                  William T. Freeman and
                  Taesung Park},
  title        = {One-Step Diffusion with Distribution Matching Distillation},
  booktitle    = {{IEEE/CVF} Conference on Computer Vision and Pattern Recognition,
                  {CVPR} 2024, Seattle, WA, USA, June 16-22, 2024},
  pages        = {6613--6623},
  year         = {2024},
  doi          = {10.1109/CVPR52733.2024.00632},
}

@inproceedings{sauer2024add,
  author       = {Axel Sauer and
                  Dominik Lorenz and
                  Andreas Blattmann and
                  Robin Rombach},
  editor       = {Ales Leonardis and
                  Elisa Ricci and
                  Stefan Roth and
                  Olga Russakovsky and
                  Torsten Sattler and
                  G{\"{u}}l Varol},
  title        = {Adversarial Diffusion Distillation},
  booktitle    = {Computer Vision - {ECCV} 2024 - 18th European Conference, Milan, Italy,
                  September 29-October 4, 2024, Proceedings, Part {LXXXVI}},
  volume       = {15144},
  pages        = {87--103},
  year         = {2024},
  doi          = {10.1007/978-3-031-73016-0\_6},
}

@article{lin2025hypir,
  author       = {Xinqi Lin and
                  Fanghua Yu and
                  Jinfan Hu and
                  Zhiyuan You and
                  Wu Shi and
                  Jimmy S. Ren and
                  Jinjin Gu and
                  Chao Dong},
  title        = {Harnessing Diffusion-Yielded Score Priors for Image Restoration},
  journal      = {{ACM} Trans. Graph.},
  volume       = {44},
  number       = {6},
  pages        = {208:1--208:21},
  year         = {2025},
  doi          = {10.1145/3763346},
}

@inproceedings{porter1984compositing,
  author       = {Thomas K. Porter and
                  Tom Duff},
  editor       = {Hank Christiansen},
  title        = {Compositing digital images},
  booktitle    = {Proceedings of the 11th Annual Conference on Computer Graphics and
                  Interactive Techniques, {SIGGRAPH} 1984, Minneapolis, Minnesota, USA,
                  July 23-27, 1984},
  pages        = {253--259},
  year         = {1984},
  doi          = {10.1145/800031.808606},
}

@article{levin2008matting,
  author       = {Anat Levin and
                  Dani Lischinski and
                  Yair Weiss},
  title        = {A Closed-Form Solution to Natural Image Matting},
  journal      = {{IEEE} Trans. Pattern Anal. Mach. Intell.},
  volume       = {30},
  number       = {2},
  pages        = {228--242},
  year         = {2008},
  doi          = {10.1109/TPAMI.2007.1177},
}

@inproceedings{xu2017deepmatting,
  author       = {Ning Xu and
                  Brian L. Price and
                  Scott Cohen and
                  Thomas S. Huang},
  title        = {Deep Image Matting},
  booktitle    = {2017 {IEEE} Conference on Computer Vision and Pattern Recognition,
                  {CVPR} 2017, Honolulu, HI, USA, July 21-26, 2017},
  pages        = {311--320},
  year         = {2017},
  doi          = {10.1109/CVPR.2017.41}
}

@inproceedings{isola2017pix2pix,
  author       = {Phillip Isola and
                  Jun{-}Yan Zhu and
                  Tinghui Zhou and
                  Alexei A. Efros},
  title        = {Image-to-Image Translation with Conditional Adversarial Networks},
  booktitle    = {2017 {IEEE} Conference on Computer Vision and Pattern Recognition,
                  {CVPR} 2017, Honolulu, HI, USA, July 21-26, 2017},
  pages        = {5967--5976},
  year         = {2017},
  doi          = {10.1109/CVPR.2017.632}
}

@article{zeng2021aotgan,
  author       = {Yanhong Zeng and
                  Jianlong Fu and
                  Hongyang Chao and
                  Baining Guo},
  title        = {Aggregated Contextual Transformations for High-Resolution Image Inpainting},
  journal      = {{IEEE} Trans. Vis. Comput. Graph.},
  volume       = {29},
  number       = {7},
  pages        = {3266--3280},
  year         = {2023},
  doi          = {10.1109/TVCG.2022.3156949}
}

@inproceedings{li2024drip,
  author       = {Xiaodi Li and
                  Zongxin Yang and
                  Ruijie Quan and
                  Yi Yang},
  title        = {{DRIP:} Unleashing Diffusion Priors for Joint Foreground and Alpha
                  Prediction in Image Matting},
  booktitle    = {Advances in Neural Information Processing Systems 38: Annual Conference
                  on Neural Information Processing Systems 2024, NeurIPS 2024, Vancouver,
                  BC, Canada, December 10 - 15, 2024},
  year         = {2024},
  url          = {http://papers.nips.cc/paper\_files/paper/2024/hash/91edff07232fb1b55a505a9e9f6c0ff3-Abstract-Conference.html}
}

@inproceedings{hu2024diffumatting,
  author       = {Xiaobin Hu and
                  Xu Peng and
                  Donghao Luo and
                  Xiaozhong Ji and
                  Jinlong Peng and
                  Zhengkai Jiang and
                  Jiangning Zhang and
                  Taisong Jin and
                  Chengjie Wang and
                  Rongrong Ji},
  title        = {DiffuMatting: Synthesizing Arbitrary Objects with Matting-Level Annotation},
  booktitle    = {Computer Vision - {ECCV} 2024 - 18th European Conference, Milan, Italy,
                  September 29-October 4, 2024, Proceedings, Part {LXVIII}},
  volume       = {15126},
  pages        = {396--413},
  year         = {2024},
  doi          = {10.1007/978-3-031-73113-6\_23}
}

@article{huang2024layerdiff,
  author       = {Lvmin Zhang and
                  Maneesh Agrawala},
  title        = {Transparent Image Layer Diffusion using Latent Transparency},
  journal      = {{ACM} Trans. Graph.},
  volume       = {43},
  number       = {4},
  pages        = {100:1--100:15},
  year         = {2024},
  doi          = {10.1145/3658150}
}

@inproceedings{hu2022lora,
  author       = {Edward J. Hu and
                  Yelong Shen and
                  Phillip Wallis and
                  Zeyuan Allen{-}Zhu and
                  Yuanzhi Li and
                  Shean Wang and
                  Lu Wang and
                  Weizhu Chen},
  title        = {LoRA: Low-Rank Adaptation of Large Language Models},
  booktitle    = {The Tenth International Conference on Learning Representations, {ICLR}
                  2022, Virtual Event, April 25-29, 2022},
  year         = {2022},
  url          = {https://openreview.net/forum?id=nZeVKeeFYf9}
}

@inproceedings{zhang2018lpips,
  author       = {Richard Zhang and
                  Phillip Isola and
                  Alexei A. Efros and
                  Eli Shechtman and
                  Oliver Wang},
  title        = {The Unreasonable Effectiveness of Deep Features as a Perceptual Metric},
  booktitle    = {2018 {IEEE} Conference on Computer Vision and Pattern Recognition,
                  {CVPR} 2018, Salt Lake City, UT, USA, June 18-22, 2018},
  pages        = {586--595},
  year         = {2018},
  doi          = {10.1109/CVPR.2018.00068}
}

@inproceedings{yu2025omnipaint,
    author    = {Yu, Yongsheng and Zeng, Ziyun and Zheng, Haitian and Luo, Jiebo},
    title     = {OmniPaint: Mastering Object-Oriented Editing via Disentangled Insertion-Removal Inpainting},
    booktitle = {Proceedings of the IEEE/CVF International Conference on Computer Vision (ICCV)},
    month     = {October},
    year      = {2025},
    pages     = {17324--17334}
}

@inproceedings{heusel2017ttur,
  author       = {Martin Heusel and
                  Hubert Ramsauer and
                  Thomas Unterthiner and
                  Bernhard Nessler and
                  Sepp Hochreiter},
  title        = {GANs Trained by a Two Time-Scale Update Rule Converge to a Local Nash
                  Equilibrium},
  booktitle    = {Advances in Neural Information Processing Systems 30: Annual Conference
                  on Neural Information Processing Systems 2017, December 4-9, 2017,
                  Long Beach, CA, {USA}},
  pages        = {6626--6637},
  year         = {2017},
  url          = {https://proceedings.neurips.cc/paper/2017/hash/8a1d694707eb0fefe65871369074926d-Abstract.html}
}

@inproceedings{jayasumana2024rethinkingfid,
  author       = {Sadeep Jayasumana and
                  Srikumar Ramalingam and
                  Andreas Veit and
                  Daniel Glasner and
                  Ayan Chakrabarti and
                  Sanjiv Kumar},
  title        = {Rethinking {FID:} Towards a Better Evaluation Metric for Image Generation},
  booktitle    = {{IEEE/CVF} Conference on Computer Vision and Pattern Recognition,
                  {CVPR} 2024, Seattle, WA, USA, June 16-22, 2024},
  pages        = {9307--9315},
  year         = {2024},
  doi          = {10.1109/CVPR52733.2024.00889}
}

@inproceedings{liu2023groundingdino,
  author       = {Shilong Liu and
                  Zhaoyang Zeng and
                  Tianhe Ren and
                  Feng Li and
                  Hao Zhang and
                  Jie Yang and
                  Qing Jiang and
                  Chunyuan Li and
                  Jianwei Yang and
                  Hang Su and
                  Jun Zhu and
                  Lei Zhang},
  title        = {Grounding {DINO:} Marrying {DINO} with Grounded Pre-training for Open-Set
                  Object Detection},
  booktitle    = {Computer Vision - {ECCV} 2024 - 18th European Conference, Milan, Italy,
                  September 29-October 4, 2024, Proceedings, Part {XLVII}},
  volume       = {15105},
  pages        = {38--55},
  year         = {2024},
  doi          = {10.1007/978-3-031-72970-6\_3}
}

@inproceedings{ravi2024sam2,
  author       = {Nikhila Ravi and
                  Valentin Gabeur and
                  Yuan{-}Ting Hu and
                  Ronghang Hu and
                  Chaitanya Ryali and
                  Tengyu Ma and
                  Haitham Khedr and
                  Roman R{\"{a}}dle and
                  Chlo{\'{e}} Rolland and
                  Laura Gustafson and
                  Eric Mintun and
                  Junting Pan and
                  Kalyan Vasudev Alwala and
                  Nicolas Carion and
                  Chao{-}Yuan Wu and
                  Ross B. Girshick and
                  Piotr Doll{\'{a}}r and
                  Christoph Feichtenhofer},
  title        = {{SAM} 2: Segment Anything in Images and Videos},
  booktitle    = {The Thirteenth International Conference on Learning Representations,
                  {ICLR} 2025, Singapore, April 24-28, 2025},
  year         = {2025},
  url          = {https://openreview.net/forum?id=Ha6RTeWMd0}
}

@inproceedings{yu2019freeform,
  author       = {Jiahui Yu and
                  Zhe Lin and
                  Jimei Yang and
                  Xiaohui Shen and
                  Xin Lu and
                  Thomas S. Huang},
  title        = {Free-Form Image Inpainting With Gated Convolution},
  booktitle    = {2019 {IEEE/CVF} International Conference on Computer Vision, {ICCV}
                  2019, Seoul, Korea (South), October 27 - November 2, 2019},
  pages        = {4470--4479},
  year         = {2019},
  doi          = {10.1109/ICCV.2019.00457}
}

@inproceedings{liu2022convnet,
  author       = {Zhuang Liu and
                  Hanzi Mao and
                  Chao{-}Yuan Wu and
                  Christoph Feichtenhofer and
                  Trevor Darrell and
                  Saining Xie},
  title        = {A ConvNet for the 2020s},
  booktitle    = {{IEEE/CVF} Conference on Computer Vision and Pattern Recognition,
                  {CVPR} 2022, New Orleans, LA, USA, June 18-24, 2022},
  pages        = {11966--11976},
  year         = {2022},
  doi          = {10.1109/CVPR52688.2022.01167}
}

@inproceedings{radford2021clip,
  author       = {Alec Radford and
                  Jong Wook Kim and
                  Chris Hallacy and
                  Aditya Ramesh and
                  Gabriel Goh and
                  Sandhini Agarwal and
                  Girish Sastry and
                  Amanda Askell and
                  Pamela Mishkin and
                  Jack Clark and
                  Gretchen Krueger and
                  Ilya Sutskever},
  editor       = {Marina Meila and
                  Tong Zhang},
  title        = {Learning Transferable Visual Models From Natural Language Supervision},
  booktitle    = {Proceedings of the 38th International Conference on Machine Learning,
                  {ICML} 2021, 18-24 July 2021, Virtual Event},
  volume       = {139},
  pages        = {8748--8763},
  year         = {2021},
  url          = {http://proceedings.mlr.press/v139/radford21a.html}
}

@inproceedings{cherti2023scaling,
  author       = {Mehdi Cherti and
                  Romain Beaumont and
                  Ross Wightman and
                  Mitchell Wortsman and
                  Gabriel Ilharco and
                  Cade Gordon and
                  Christoph Schuhmann and
                  Ludwig Schmidt and
                  Jenia Jitsev},
  title        = {Reproducible Scaling Laws for Contrastive Language-Image Learning},
  booktitle    = {{IEEE/CVF} Conference on Computer Vision and Pattern Recognition,
                  {CVPR} 2023, Vancouver, BC, Canada, June 17-24, 2023},
  pages        = {2818--2829},
  year         = {2023},
  doi          = {10.1109/CVPR52729.2023.00276}
}

@inproceedings{mescheder2018gan,
  author       = {Lars M. Mescheder and
                  Andreas Geiger and
                  Sebastian Nowozin},
  editor       = {Jennifer G. Dy and
                  Andreas Krause},
  title        = {Which Training Methods for GANs do actually Converge?},
  booktitle    = {Proceedings of the 35th International Conference on Machine Learning,
                  {ICML} 2018, Stockholmsm{\"{a}}ssan, Stockholm, Sweden, July
                  10-15, 2018},
  volume       = {80},
  pages        = {3478--3487},
  year         = {2018},
  url          = {http://proceedings.mlr.press/v80/mescheder18a.html}
}
